\documentclass{article}
\usepackage{log_2023}						

\usepackage{microtype}
\usepackage{graphicx}
\usepackage{booktabs} 

\usepackage{amssymb,bm,amsthm}
\usepackage{enumerate}
\usepackage{color, colortbl}
\usepackage{booktabs, multirow}
\definecolor{darkgreen}{rgb}{0, 0.5, 0}
\definecolor{red}{rgb}{1, 0, 0}
\definecolor{purple}{rgb}{0.5, 0, 0.5}
\usepackage{chngpage}
\usepackage{comment}
\usepackage{mfirstuc}
\usepackage{mathtools}
\usepackage{lettrine}

\newcommand\ie{\textit{i.e.,\xspace}}

\newcommand\wrt{\textit{w.r.t.}}
\newcommand\etc{\textit{etc.}}

\newcommand{\red}{\textcolor{red}}

\newcommand{\beq}{\begin{equation}}
\newcommand{\eeq}{\end{equation}}
\newcommand{\beqnn}{\begin{equation*}}
\newcommand{\eeqnn}{\end{equation*}}
\newcommand{\beqy}{\begin{eqnarray}}
\newcommand{\eeqy}{\end{eqnarray}}
\newcommand{\beqynn}{\begin{eqnarray*}}
\newcommand{\eeqynn}{\end{eqnarray*}}
\newcommand{\bit}{\begin{itemize}}
\newcommand{\eit}{\end{itemize}}
\newcommand{\ben}{\begin{enumerate}}
\newcommand{\een}{\end{enumerate}}
\newcommand{\bex}{\begin{example}}
\newcommand{\eex}{\end{example}}

\newcommand{\bary}{\begin{array}}
\newcommand{\eary}{\end{array}}
\newcommand{\bmx}{\begin{bmatrix}}
\newcommand{\emx}{\end{bmatrix}}
\newcommand{\bsmx}{\left[\begin{smallmatrix}}
\newcommand{\esmx}{\end{smallmatrix}\right]}
\newcommand{\bmxc}[1]{\left[\begin{array}{@{}#1@{}}}
\newcommand{\emxc}{\end{array}\right]}
\newcommand{\bcn}{\begin{center}}
\newcommand{\ecn}{\end{center}}

\usepackage{amsmath,amsfonts,bm}

\def\1{\bm{1}}

\def\vv{{\bm{v}}}
\def\vw{{\bm{w}}}

\DeclareMathAlphabet{\mathsfit}{\encodingdefault}{\sfdefault}{m}{sl}
\SetMathAlphabet{\mathsfit}{bold}{\encodingdefault}{\sfdefault}{bx}{n}

\newcommand{\KL}{D_{\mathrm{KL}}}

\usepackage{subfig}
\usepackage{wrapfig}
\usepackage{xspace}
\usepackage{multirow, bigstrut}
\usepackage{amsmath}
\usepackage{bbm}
\usepackage{algorithmic}
\usepackage{algorithm}
\usepackage{wrapfig}

\usepackage[numbers,compress,sort]{natbib}	

\title[MUDiff: Unified Diffusion for Complete Molecule Generation]{MUDiff: Unified Diffusion for Complete Molecule Generation}

\author[C. Hua et al.]{%
Chenqing Hua\\
McGill University; Mila\\
\email{chenqing.hua@mail.mcgill.ca}\And
Sitao Luan\\
McGill University; Mila\\
\email{sitao.luan@mail.mcgill.ca}\And
Minkai Xu\\
Stanford University\\
\email{minkai@cs.stanford.edu}\And
Rex Ying\\
Yale University\\
\email{rex.ying@yale.edu}\And
Jie Fu\thanks{Corresponding to Jie Fu: jiefu@ust.hk}\\
HKUST; Mila\\
\email{jiefu@ust.hk}\And
Stefano Ermon\\
Stanford University\\
\email{ermon@cs.stanford.edu}\And
Doina Precup\\
McGill University; Mila\\
\email{dprecup@cs.mcgill.ca}
}

\begin{document}

\maketitle

\begin{abstract}
Molecule generation is a very important practical problem, with uses in drug discovery and material design, and AI methods promise to provide useful solutions. However, existing methods for molecule generation focus either on 2D graph structure or on 3D geometric structure, which is not sufficient to represent a complete molecule as 2D graph captures mainly topology while 3D geometry captures mainly spatial atom arrangements. Combining these representations is essential to better represent a molecule. In this paper, we present a new model for generating a comprehensive representation of molecules, including atom features, 2D discrete molecule structures, and 3D continuous molecule coordinates, by combining discrete and continuous diffusion processes. The use of diffusion processes allows for capturing the probabilistic nature of molecular processes and exploring the effect of different factors on molecular structures. Additionally, we propose a novel graph transformer architecture to denoise the diffusion process. The transformer adheres to 3D roto-translation equivariance constraints, allowing it to learn invariant atom and edge representations while preserving the equivariance of atom coordinates. This transformer can be used to learn molecular representations robust to geometric transformations. We evaluate the performance of our model through experiments and comparisons with existing methods, showing its ability to generate more stable and valid molecules. Our model is a promising approach for designing stable and diverse molecules and can be applied to a wide range of tasks in molecular modeling. Our codes and models are available on \red{\url{https://github.com/WillHua127/mudiff}}
\end{abstract}

\vspace{-0.2cm}
\section{Introduction}
\vspace{-0.2cm}
\label{sec:introduction}
Generative models for molecules in machine learning have gained significant attention in recent years as a promising approach for the design and discovery of novel molecules with desired properties \cite{kipf2016variational, satorras2021n, xu2022geodiff}. These models are trained on a dataset of known molecular structures and can then generate unseen molecules similar to those in the training dataset.
As one specific type of generative models, diffusion models are based on the idea of learning small changes in the molecular structures \cite{ hoogeboom2022equivariant}. The model learns the likelihood of these changes and can generate new molecules by sampling from the learned distribution. This approach has been used to generate new drug candidates, optimize drug properties, \etc \cite{jo2022score, hoogeboom2022equivariant, vignac2022digress}.

While 2D graph structures capture the topology and connectivity of molecules \cite{duvenaud2015convolutional, gilmer2017neural}, 3D geometric structures provide an insight into the spatial arrangements of atoms \cite{schutt2017schnet, xu2022geodiff}, the two structural information are essential for a comprehensive representation of a molecule. 
So, learning 2D and 3D structures together leads to an accurate and complete molecule representation.
However, existing generative models for molecules focus solely on either 2D or 3D molecular data generation \cite{satorras2021n, hoogeboom2022equivariant, vignac2022digress}, thus limiting their ability to exploit an accurate and complete representation of molecules. This limitation highlights the need for joint generation and learning of 2D and 3D molecular data, and motivates us to work on the joint generation problem. To address the gap, we are motivated to propose a novel generative model that jointly generates 2D and 3D molecular data, capturing both the topological information from 2D graphs, and spatial atom arrangements from 3D geometry, thereby enabling a more holistic understanding of molecular structures.

In this paper, we present a novel approach to overcome the aforementioned limitations by jointly generating 2D and 3D aspects of molecules, yielding a complete representation of molecules by learning the graph connectivity and atom arrangements together.
We propose a diffusion generative model that co-generates the 2D graph structure and 3D geometric structure of a molecule, named MUDiff, and a transformer model that co-learns both molecular structures, named MUformer. 
Our diffusion model simultaneously adds continuous noises to the continuous features, including atom features and coordinates, and discrete noises to the categorical features, including the graph structure. The denoising model then makes predictions for the clean graph structure, as well as estimations of noises for atom features and coordinates,
The primary innovation of our denoising transformer model lies in designing an attention mechanism that facilitates interaction between 2D and 3D structural information. This design enables our transformer to concurrently compute graph connectivity and spatial atom arrangements. Moreover, when computing 3D geometric structures, the denoising transformer model adheres to the critical 3D roto-translation equivariance constraints. This compliance ensures insensitivity to geometric transformations on the molecule, allowing the entire diffusion process to adhere to these constraints. Through the novel designs, our model can generate and learn a comprehensive molecular representation that captures both 2D and 3D structures, addressing the aforementioned limitations.

Furthermore, a distinct advantage of MUDiff and MUformer is the ability to function independently when either 2D or 3D structural information is missing. Our model remains effective in generating and learning a complete representation of molecules, even when the input data lacks the 2D graph structure or the 3D geometric structure. This design is particularly useful because datasets sometimes have missing 3D coordinates and geometry, resulted from limitations in experimental techniques or the unavailability of suitable computational resources \cite{mobley2017predicting}. For instance, the conformational analysis of molecules, which involves determining the 3D structures that result from the rotation of single bonds, is of critical importance in understanding molecular interactions \cite{leach2011molecular, copeland2013evaluation}. However, obtaining accurate conformational data can be computationally demanding \cite{gaulton2012chembl}. In such cases, MUDiff and MUformer can provide a robust and versatile solution for handling incomplete molecular datasets, ensuring comprehensive molecular representations even when faced with such challenges.

Emperically, MUDiff generates 7.9\% more stable molecules and increases molecular uniqueness by ~2\% compared to existing methods (see Sec~\ref{sec:molecule.generation}). 
Additionally, even when trained with limited 3D structures, MUDiff still achieve competitive performance compared to existing methods trained with complete 3D structures (see Sec~\ref{sec:molecule.generation.limit.3d}). The results show the better model performance in generating stable and valid molecules, emphasizing the importance of joint 2D and 3D molecule generation for advancing the field.
In Sec~\ref{sec:diffusion.process}, we present the molecule unified diffusion model (MUDiff), followed by the introduction of the molecule unified transformer model (MUformer) in Sec~\ref{sec:graph.transformer} and App~\ref{app:muformer}. The MUformer architecture is visually demonstrated in Fig~\ref{fig:mudiff}. Detailed experimental results can be found in Sec~\ref{sec:experiments} and App~\ref{app:additional.experiment}. 

\vspace{-0.5cm}
\section{Preliminaries}
\vspace{-0.2cm}
\label{sec:prelimiary.notation}
\subsection{Diffusion Models}
\vspace{-0.3cm}
\label{sec:diffusion.processes}
Diffusion models consist of a noising model and a denoising network. The noising model $q$ adds noise to a data point $\mathbf{X}$ to generate a sequence of noisy points $\{\mathbf{\Tilde{X}}_t\}^{T}_{t=0}$. This process follows the Markov property, $q(\mathbf{\Tilde{X}}_0,\ldots,\mathbf{\Tilde{X}}_T | \mathbf{X})=q(\mathbf{\Tilde{X}}_0|\mathbf{X})\prod_{t=1}^T q(\mathbf{\Tilde{X}}_t|\mathbf{\Tilde{X}}_{t-1})$.
The denoising network $\psi_{\theta}$ aims to reverse the noising process: given a noisy point $\mathbf{\Tilde{X}}_t$, it predicts a clean estimate $\mathbf{\hat{X}}=\psi_{\theta}(\mathbf{\Tilde{X}}_t, t)$ of $\mathbf{X}$.

\vspace{-0.1cm}
\noindent \textbf{Continuous Data} \quad
The \textit{noising process} in diffusion models for continuous data point $\mathbf{X}$ can be represented by a multivariate normal distribution,
$
    q(\mathbf{\Tilde{X}}_t|\mathbf{X}) = \mathcal{N}(\mathbf{\Tilde{X}}_t|\alpha_t\mathbf{X}, \sigma^2_t\mathbf{I}),
$
where $\alpha_t\in \mathbb{R}^+$ controls the amount of signal retained and $\sigma_t\in \mathbb{R}^+$ represents the amount of Gaussian noise added, and $\alpha_t$ smoothly transitions from $ 1$ to $0$. Following \cite{ho2020denoising}, we choose $\alpha_t = \sqrt{1-\sigma^2_t}$ in order to obtain a variance preserving process, and the signal-to-noise ratio SNR$(t) = {\alpha^2_t}/{\sigma^2_t}$ is defined by \cite{kingma2021variational}.
For every two time steps $t, t-1$, the noising process is
$
q(\mathbf{\Tilde{X}}_t|\mathbf{\Tilde{X}}_{t-1}) = \mathcal{N}(\mathbf{\Tilde{X}}_t|\alpha_{t|t-1}\mathbf{\Tilde{X}}_{t-1}, \sigma^2_{t|t-1}\mathbf{I}),
$
where $\alpha_{t|t-1}=\alpha_t/\alpha_{t-1}$ and $\sigma^2_{t|t-1} = \sigma^2_t - \alpha_{t|t-1}^2\sigma^2_{t-1}$.

The posterior of the transitions gives the \textit{denoising process},
$
    q(\mathbf{\hat{X}}_{t-1}|\mathbf{\hat{X}}_{t}, \mathbf{{X}}) = \mathcal{N}(\mathbf{\hat{X}}_{t-1}|\bm{\mu}_{t\to t-1}(\mathbf{\Tilde{X}}_t, \mathbf{{X}}), \sigma^2_{t\to t-1}\mathbf{I}),
$
where the functions are defined as $\bm{\mu}_{t\to t-1}(\mathbf{\Tilde{X}}_t, \mathbf{{X}})=\frac{\alpha_{t|t-1}\sigma^2_{t-1}}{\sigma^2_t}\mathbf{\Tilde{X}}_t + \frac{\alpha_{t-1}\sigma^2_{t|t-1}}{\sigma^2_t}\mathbf{{X}}, \sigma_{t\to t-1}=\frac{\sigma_{t|t-1}\sigma_{t-1}}{\sigma_t}$. 
In the true denoising process, the clean data $\mathbf{{X}}$ is unknown, so it is replaced by the network approximation $\mathbf{\hat{X}}$ as,
\begin{equation}
\label{eq:posterior.atom}
    p(\mathbf{\hat{X}}_{t-1}|\mathbf{\hat{X}}_{t}) = \mathcal{N}(\mathbf{\hat{X}}_{t-1}|\bm{\mu}_{t\to t-1}(\mathbf{\Tilde{X}}_t, \mathbf{\hat{X}}), \sigma^2_{t\to t-1}\mathbf{I}).
\end{equation}

\vspace{-0.4cm}
\noindent \textbf{Discrete Data} \quad
Discrete objects, such as graph structures, may not be well suited for Gaussian noise models as they can destroy sparsity and connectivity, as suggested by \cite{vignac2022digress}.
Following \cite{austin2021structured, vignac2022digress}, we use a series of transition matrices $\{Q_t\}_{t=0}^T$ to represent noise on one-hot encoded discrete data points, where ${Q}_{{t}_{{ij}}}=q(\mathbf{X}_t=j|\mathbf{X}_{t-1}=i)$ 
is the probability of transitioning from state $i$ to state $j$.
We obtain noisy data by multiplying the clean data point with a transition matrix, 
$
        q(\mathbf{\Tilde{X}}_t | \mathbf{\Tilde{X}}_{t-1}) = \mathbf{\Tilde{X}}_tQ_{t}, \ q(\mathbf{\Tilde{X}}_t | \mathbf{X}) = \mathbf{X}\bar{Q}_{t},
$
where $\bar{Q}_t = Q_tQ_{t-1}\ldots Q_0$.
The posterior distribution is computed using the Bayes rule,
$
    q(\mathbf{\Tilde{X}}_{t-1}| \mathbf{\Tilde{X}}_{t}, \mathbf{{X}}) \propto \mathbf{\Tilde{X}}_{t} Q_t^T \odot \mathbf{X}\bar{Q}_{t-1},
$
where $Q^T$ represents the transpose of the transition matrix $Q$, and $\odot$ denotes the Hadamard product.

\vspace{-0.4cm}
\subsection{The Basics of Molecules}
\vspace{-0.2cm}
\label{sec:basics}
A molecule is a group of atoms held together by chemical bonds, which can be classified into various types based on the nature of the bond. The structure of a molecule can be visualized and represented in both 2D and 3D forms, with the 2D representation showing the connectivity of the atoms and the 3D representation showing the arrangement of the atoms in space. To completely describe a molecule, we represent it as a tuple $\mathbf{M}=(\mathbf{H, E, X})$, where $\mathbf{H} \in \mathbb{R}^{n \times d}$ denotes the collection of atoms, $n$ is the number of atoms, and $d$ is the feature dimension; $\mathbf{E} \in \mathbb{R}^{n \times n \times b}$ is the 2D graph representation for chemical bonds, the bond type is represented by one-hot encoding, and $b$ is the number of bond(edge) types; $\mathbf{X} \in \mathbb{R}^{n \times 3}$ represents the 3D geometric structure and each row indicates the position of the atom in the Euclidean space. For simplicity, we assume that molecules are fully connected and include \textit{no-bond} as a special edge type.  To account for symmetry, we use symmetric edge representations, \ie{} $\mathbf{E}=\mathbf{E}^T$. 

\vspace{-0.1cm}
\noindent \textbf{3D Molecule Representation}  \quad
For each pair of atoms $(i,j)$ in the 3D space, we process the Euclidean distance between them using an exponential normal radial basis function \cite{schutt2017schnet}, \ie{} $f_{\text{RBF}^k}(d_{ij})=\exp{(-\beta_k(\exp{(-d_{ij})-\mu_k})^2)}$, where $k$ is the number of basis kernels, $d_{ij}$ is the distance between atoms $i$ and $j$, $\beta_k$ and $\mu_k$ are fixed parameters determining the function's center, and width, respectively. These parameters are initialized as per \cite{unke2019physnet}.

To smooth out the transition to $0$ as the distance $d_{ij}$ approaches a cutoff distance of $d_\text{cut}=5$\r{A}, we also apply a cosine cutoff function $f_{\cos}(d_{ij})$, \ie{} $f_{\cos}(d_{ij})=\frac{1}{2}(\cos{(\frac{\pi d_{ij}}{d_{\text{cut}}})}+1)$ if $d_{ij} \leq d_{\text{cut}}$, and $f_{\cos}(d_{ij})=0$ if $d_{ij} > d_{\text{cut}}$. In the model, we will use both $f_{\cos}(d_{ij})$ and  $f_{\cos}(f_{\text{RBF}^k}(d_{ij}))$.

\vspace{-0.4cm}
\section{MUDiff: \underline{M}olecule \underline{U}nified \underline{Diff}usion}
\vspace{-0.2cm}
\label{sec:diffusion.process}
Both the continuous and discrete elements of molecules are essential in order to depict a comprehensive molecular representation, however, the existing models \cite{satorras2021n, xu2022geodiff, hoogeboom2022equivariant, vignac2022digress} have only been able to generate a portion of these components.
Our diffusion model is designed to denoise continuous and discrete aspects of a molecule separately. The continuous aspects encompass atom features and 3D coordinates, while the discrete aspects include molecular structure. This separation allows for independent handling of atoms and edges, a similar approach is shown to be successful in image diffusion models \cite{austin2021structured}, but unexplored for molecules. By jointly generating the continuous 3D geometry and discrete 2D graph representation, our model enhances the representation of atom and edge features, yielding a more comprehensive and holistic understanding of the molecule that incorporates both geometric and topological information.

\vspace{-0.4cm}
\subsection{Diffusion Process}
\vspace{-0.2cm}
Our diffusion model distinctly applies noises to atom and edge representations to enhance the generative process. Specifically, we apply continuous noises to atom representations, encompassing both atom features and coordinates, while introducing discrete noises to edge representations, which correspond to the graph structure. This targeted approach differentiates our diffusion process from previous molecule diffusion models, allowing for a more comprehensive generation of molecules that captures both geometric and topological information.

\vspace{-0.1cm}
\noindent \textbf{Atom Features and Coordinates} \quad
As introduced in Sec~\ref{sec:diffusion.processes}, we add Gaussian noise to atom features and coordinates at each time step $t$, with ${\bm{\epsilon}}_{\mathbf{H}}^t\sim \mathcal{N}_{\mathbf{H}}(\mathbf{0, I}) \in \mathbb{R}^{n\times d}, {\bm{\epsilon}}_{\mathbf{X}}^t\sim \mathcal{N}_{\mathbf{X}}(\mathbf{0, I}) \in \mathbb{R}^{n\times 3}$, where $n$ is the number of atoms, $d$ denotes the feature dimension. For the 3D coordinates, we follow \cite{kohler2020equivariant} to use the linear subspace of zero center of gravity for ${\mathcal{N}_{\mathbf{X}}}$ such that $\sum_{i}\mathbf{x}_i=0$. This leads to noisy atom features and coordinates,
\begin{equation}
\label{eq:noisy.atom.features}
    \begin{aligned}
        &\mathbf{\tilde{H}}_t = \alpha_t \mathbf{H} + \sigma_t {\bm{\epsilon}}^t_{\mathbf{H}},\; \mathbf{\tilde{X}}_t = \alpha_t \mathbf{X} + \sigma_t {\bm{\epsilon}}^t_{\mathbf{X}}.
    \end{aligned}
\end{equation}
This method ensures that the perturbations applied to the 3D coordinates do not affect the center of gravity of the molecule, allowing for the denoising process to be invariant \wrt{} to translations.

\vspace{-0.1cm}
\noindent \textbf{Edge Features} \quad
Following Sec~\ref{sec:diffusion.processes}, we transform the discrete clean edge type to obtain noisy ones,
\begin{equation}
    \mathbf{\tilde{E}}_t=\mathbf{E}\bar{Q}_t.
\end{equation}
where the transition matrix $\bar{Q}_t$ is obtained by $\bar{Q}_t = \alpha_t \mathbf{I} + (1-\alpha_t){\mathbbm{1}_b\mathbbm{1}^t_b}/b \in \mathbb{R}^{b\times b}$. We use uniform transitions over the number of edge types $b$ \cite{austin2021structured, vignac2022digress}, resulting in a uniform limit distribution $q_{\infty}$ over edge categories (see App~\ref{app:limit.edge}). Additionally, since molecules are always undirected graphs, we only apply noise to the upper triangular of the edge representation matrix and then symmetrize the matrix, which ensures that changes made on the edges are consistent across the graph.

\vspace{-0.4cm}
\subsection{Denoising Process}
\vspace{-0.2cm}
To date, no existing models can simultaneously predict the features of atoms $\mathbf{H}$, their coordinates $\mathbf{X}$, and the structures of molecules $\mathbf{E}$. To address this gap, we introduce a novel denoising network, named MUformer, which learns the denoising process to make predictions for the comprehensive representation of molecules. This model is unique in its ability to consider all aspects of the molecule in a unified manner while ensuring that the denoising process is equivariant, as suggested by \cite{xu2022geodiff}.

\begin{wrapfigure}{R}{0.46\textwidth}
\vspace{-0.7cm}
\begin{minipage}{0.46\textwidth}
\vspace{-0.3cm}
  \begin{algorithm}[H]
    \caption{Training MUDiff}
    \label{algo:training}
    \footnotesize
    \begin{algorithmic}[1]
        \STATE \textbf{Input:} A complete molecule $\mathbf{M=(H,E,X)}$
        \STATE Sample $t\sim \mathcal{U}(1,\cdots,T)$ 
        \STATE Sample ${\bm{\epsilon}}_\mathbf{H}, {\bm{\epsilon}}_\mathbf{X} \sim \mathcal{N}(\mathbf{0, I})$ 
        \STATE Subtract center of gravity from ${\bm{\epsilon}}_\mathbf{X}$ 
        \STATE Compute $\mathbf{\tilde{H}}_t = \alpha_t \mathbf{H} + \sigma_t {\bm{\epsilon}}^t_{\mathbf{H}}, \ \mathbf{\tilde{X}}_t = \alpha_t \mathbf{X} + \sigma_t {\bm{\epsilon}}^t_{\mathbf{X}}$
        \STATE Sample $\mathbf{\Tilde{E}}_t \sim \mathbf{E}\Tilde{Q}_t$ 
        \STATE Compute $\mathbf{\hat{{\bm{\epsilon}}}}^t_{\mathbf{H}}, \mathbf{\hat{{\bm{\epsilon}}}}^t_{\mathbf{X}}, p(\mathbf{\hat{E}}) =
        \psi_{\theta}([\mathbf{\tilde{H}}_t, \frac{t}{T}], \mathbf{\tilde{X}}_t, \mathbf{\tilde{E}}_t) - (\mathbf{0}, \mathbf{\tilde{X}}_t, \mathbf{0})$ 
        \STATE Minimize $\|\mathbf{{{\bm{\epsilon}}}}^t_{\mathbf{H}}-\mathbf{\hat{{\bm{\epsilon}}}}^t_{\mathbf{H}}\|^2 + \|\mathbf{{{\bm{\epsilon}}}}^t_{\mathbf{X}}-\mathbf{\hat{{\bm{\epsilon}}}}^t_{\mathbf{X}}\|^2+\text{CE}(\mathbf{{E}}, p(\mathbf{\hat{E}}))$
    \end{algorithmic}
\end{algorithm}
\end{minipage}
\vspace{-0.7cm}
\end{wrapfigure}
The denoising network, denoted by $\psi_{\theta}$,  takes as input a noisy molecule $\mathbf{\tilde{M}}_t = (\mathbf{\tilde{H}}_t, \mathbf{\tilde{E}}_t, \mathbf{\tilde{X}}_t)$ and outputs estimates for the clean molecule $\mathbf{\hat{M}}$. The detailed architecture and methodology of MUformer are presented in Sec~\ref{sec:graph.transformer}, showcasing its capacity to generate comprehensive molecular representations that encompass atom features, coordinates, and molecular structures.

\vspace{-0.1cm}
\noindent \textbf{Network Estimation} \quad
Instead of directly predicting the atom representations $\mathbf{\hat{H}, \hat{X}}$, the network attempts to predict the Gaussian noises for atom features and coordinates $\mathbf{\hat{{\bm{\epsilon}}}}_{\mathbf{H}}, \mathbf{\hat{{\bm{\epsilon}}}}_{\mathbf{X}}$, as it has been shown to be easier to optimize in \cite{ho2020denoising}. This approach allows the network to differentiate between the noise added by the noising process and the ground-truth representations, $\mathbf{{H}},\mathbf{{X}}$. The network takes as input a noisy molecule, where atom features are concatenated with the normalized time step $\frac{t}{T}$, and predicts the probability of edge features, as well as the estimates of noises for atom features and coordinates,
\begin{equation}
\label{eq:uncond}
\mathbf{\hat{{\bm{\epsilon}}}}^t_{\mathbf{H}}, \mathbf{\hat{{\bm{\epsilon}}}}^t_{\mathbf{X}}, p(\mathbf{\hat{E}})=
\psi_{\theta}([\mathbf{\tilde{H}}_t, \frac{t}{T}], \mathbf{\tilde{X}}_t, \mathbf{\tilde{E}}_t) - (\mathbf{0}, \mathbf{\tilde{X}}_t, \mathbf{0}),
\end{equation}
where the input coordinates are then subtracted from the estimated noise for coordinates to ensure that the outputs lie on the zero center of gravity subspace, as suggested by \cite{hoogeboom2022equivariant}. We subsequently obtain estimates of atom features and coordinates by
\begin{equation}
\label{eq.network.estimate.h.x}
\begin{aligned}
    & \mathbf{\hat{H}} = \frac{1}{\alpha_t} \mathbf{\Tilde{H}}_t - \frac{\sigma_t}{\alpha_t}\mathbf{\hat{{\bm{\epsilon}}}}^t_{\mathbf{H}} ,\; \mathbf{\hat{X}} = \frac{1}{\alpha_t} \mathbf{\Tilde{X}}_t - \frac{\sigma_t}{\alpha_t}\mathbf{\hat{{\bm{\epsilon}}}}^t_{\mathbf{X}}.
\end{aligned}
\end{equation}

\vspace{-0.1cm}
\noindent \textbf{Training Objective} \quad
For atom features and coordinates, the objective is to accurately predict the true noise present in the observations of atom features and coordinates.
To achieve this, we follow the approach outlined in \cite{hoogeboom2022equivariant} and minimize the distance between the true noise and the estimates of noise predicted by the network $\psi_\theta$. The objectives for atoms are defined as,
\begin{equation}
\begin{aligned}
\label{eq:training.atom.loss}
    \mathcal{L}^{\mathbf{H}}_{{t}} &= \frac{1}{2} \mathbb{E}_{{\bm{\epsilon}}_{\mathbf{H}}^t \sim N_{\mathbf{H}}(\mathbf{0,I})}\left[ \omega(t) \| {\bm{\epsilon}}_{\mathbf{H}}^t-\mathbf{\hat{{\bm{\epsilon}}}}^t_{\mathbf{H}} \|^2\right], \
    \mathcal{L}^{\mathbf{X}}_{{t}} &= \frac{1}{2} \mathbb{E}_{{\bm{\epsilon}}_{\mathbf{X}}^t \sim N_{\mathbf{X}}(\mathbf{0,I})}\left[\omega(t) \| {\bm{\epsilon}}_{\mathbf{X}}^t-\mathbf{\hat{{\bm{\epsilon}}}}^t_{\mathbf{X}} \|^2\right],
\end{aligned}
\end{equation}
where $\omega(t) = (1 - \text{SNR}(t-1)/\text{SNR}(t))$. To stabilize the training process, we set $\omega(t)=1$ during the training phase, as suggested by \cite{ho2020denoising, hoogeboom2022equivariant}.

To handle edge features, we approach it as a classification problem and minimize the cross-entropy loss for each atom pair $(i,j)\in \mathbf{E}$. The loss is calculated between the actual edge type and the predicted edge probability distribution,
\begin{equation}
\label{eq:training.edge.loss}
    \mathcal{L}_t^{\mathbf{E}} = \mathbb{E}_{(i,j) \sim \mathbf{E}}\left[\mathbf{E}_{ij} \log(p(\mathbf{\hat{E}}_{ij}))\right].
\end{equation}
At every time step $t$, the total loss is computed as the sum of the three losses,
$\mathcal{L}_t =  \mathcal{L}^{\mathbf{H}}_{{t}} +  \mathcal{L}^{\mathbf{E}}_{{t}} + \mathcal{L}^{\mathbf{X}}_{{t}}$.

\begin{wrapfigure}{R}{0.46\textwidth}
\vspace{-0.5cm}
    \begin{minipage}{0.46\textwidth}
    \vspace{-0.3cm}
      \begin{algorithm}[H]
    \centering
    \caption{Sampling from MUDiff}
    \label{algo:sampling}
    \footnotesize
    \begin{algorithmic}[1]
        \STATE Sample $\mathbf{\Tilde{M}}_T$: $\mathbf{\Tilde{H}}_T, \mathbf{\Tilde{X}}_T \sim \mathcal{N}(\mathbf{0, I}), \ \mathbf{\Tilde{E}}_T \sim q_{\infty}$
        \FOR{$t=T,T-1,\ldots,1$}
        \STATE Compute $\mathbf{\hat{{\bm{\epsilon}}}}^t_{\mathbf{H}}, \mathbf{\hat{{\bm{\epsilon}}}}^t_{\mathbf{X}}, \mathbf{\hat{E}} =
        \psi_{\theta}([\mathbf{\tilde{H}}_t, \frac{t}{T}], \mathbf{\tilde{X}}_t, \mathbf{\tilde{E}}_t) - (\mathbf{0}, \mathbf{\tilde{X}}_t, \mathbf{0})$ 
        \STATE Sample $\mathbf{\Tilde{E}}_{t-1} \sim p(\mathbf{\Tilde{E}}_{t-1}|\mathbf{\Tilde{E}}_t)$
        \STATE Sample ${\bm{\epsilon}}_{\mathbf{H}}, {\bm{\epsilon}}_{\mathbf{X}}\sim \mathcal{N}(\mathbf{0, I})$
        \STATE Compute $\mathbf{\Tilde{H}}_{t-1}=\frac{\mathbf{\Tilde{H}}_{t}}{\alpha_{t|t-1}}-\frac{\sigma^2_{t|t-1} \mathbf{\hat{{\bm{\epsilon}}}}^t_{\mathbf{H}}}{\alpha_{t|t-1} \sigma_t} + \sigma_{t\to t-1}\bm{\epsilon}_{\mathbf{H}}$
        \STATE Subtract center of gravity from ${\bm{\epsilon}}_\mathbf{X}$ 
        \STATE Compute $\mathbf{\Tilde{X}}_{t-1}=\frac{\mathbf{\Tilde{X}}_{t}}{\alpha_{t|t-1}}-\frac{\sigma^2_{t|t-1} \mathbf{\hat{{\bm{\epsilon}}}}^t_{\mathbf{X}}}{\alpha_{t|t-1} \sigma_t} + \sigma_{t\to t-1}\bm{\epsilon}_{\mathbf{X}}$
        \ENDFOR
        \STATE Sample $\mathbf{M}\sim p(\mathbf{M}|\mathbf{\Tilde{M}}_0)$
    \end{algorithmic}
\end{algorithm}
\end{minipage}
\vspace{-0.6cm}
\end{wrapfigure}
The entire training process is described in Algorithm~\ref{algo:training}. Additionally, the derivation of the variational evidence lower bound on the likelihood can be found in App~\ref{app:lower.bound}.

\vspace{-0.1cm}
\noindent \textbf{Sampling} \quad
Once the model is trained, it can be used to sample new molecules. 
The true sampling process $p(\mathbf{\Tilde{M}}_{t-1}|\mathbf{\Tilde{M}}_{t})$ uses the approximation of a complete molecule $\mathbf{\hat{M} = (\hat{H}, \hat{E}, \hat{X})}$.
The complete molecule is sampled by taking the product of the posterior distributions of atom features, coordinates, and edge features as 
\begin{equation}
    p(\mathbf{\Tilde{M}}_{t-1}|\mathbf{\Tilde{M}}_{t}) = p(\mathbf{\Tilde{H}}_{t-1} | \mathbf{\Tilde{H}}_{t})
    p(\mathbf{\Tilde{E}}_{t-1} | \mathbf{\Tilde{E}}_{t})
    p(\mathbf{\Tilde{X}}_{t-1} | \mathbf{\Tilde{X}}_{t}),
\end{equation}
with the posterior distributions of atom features and coordinates from Eq~\ref{eq:posterior.atom} and the posterior distribution of edges defined in the App~\ref{app:posterior.edge}. 
The sampling process is described in Algorithm~\ref{algo:sampling}. Also, the zeroth likelihood estimation is explained in App~\ref{app:zeroth.likelihood}.

\vspace{-0.5cm}
\section{MUformer: \underline{M}olecule \underline{U}nified Trans\underline{former}}
\vspace{-0.3cm}
\label{sec:graph.transformer}
To learn the complete molecular representation, in this section, we propose a novel equivariant graph transformer MUformer, denoted by $\psi_{\theta}$ (visualized in Fig~\ref{fig:mudiff}), which contains 6 encoding functions (Sec~\ref{sec:encodings}\&App~\ref{app:muformer.encoding}), 4 attention biases (Sec~\ref{sec:attention_biases}\&App~\ref{app:transformer.arch}), and 2 attention channels (Sec~\ref{sec:transformer_channels}\&App~\ref{app:transformer.arch}). It takes as input a complete molecule $\mathbf{M} = (\mathbf{H, E, X})$ with $\mathbf{H} \in \mathbb{R}^{n \times d}, \mathbf{E} \in \mathbb{R}^{n \times n \times b}, \mathbf{X} \in \mathbb{R}^{n \times 3}$, and outputs the predicted molecule $\mathbf{\hat{M}} = (\mathbf{\hat{H}, \hat{E}, \hat{X}})$. For clarity, we refer to the 2D molecular structure as $\mathbf{M}^{\text{2D}}=(\mathbf{H}, \mathbf{E})$, and the 3D geometric structure as $\mathbf{M}^{\text{3D}}=(\mathbf{H}, \mathbf{X})$. In the following subsections, we will introduce each component of our MUformer in details.

Our MUformer architecture can be used under different input conditions. When only 2D molecular information is available, only the invariant channel is activated, and the model makes predictions for atom and edge features only. When only 3D molecular information is available, only the equivariant channel is activated, and the model makes predictions for atom features and coordinates only. When both 2D and 3D molecular data are provided, the invariant and equivariant channels are activated, and the model can make predictions for the complete molecule, including atom features, molecular structure, and geometric structure. A detailed introduction of MUformer is discussed in App~\ref{app:muformer}.

\vspace{-0.4cm}
\subsection{Encodings}
\vspace{-0.2cm}
\label{sec:encodings}
The MUformer consists of 6 encoding functions, with 3 being message-passing based, designed to incorporate atomic, positional, and structural information into a concise and expressive representation, particularly suited for handling graph-structured inputs. A more complete introduction of MUformer encoding functions are discussed in App~\ref{app:muformer.encoding}.

\vspace{-0.2cm}
\noindent \textbf{1. Atom Encoding} \quad
The authors in \cite{ying2021transformers} propose a method of utilizing in-degree $\text{deg}^-$ and out-degree $\text{deg}^+$ obtained from 2D molecular graphs $\mathbf{M}^{\text{2D}}$ to incorporate centrality information into the atom-wise encoding process, $\mathbf{z}^{\text{1D}}_{\mathbf{h}_i}$ for node $i$ is,
\begin{equation}
    \mathbf{z}^{\text{1D}}_{\mathbf{h}_i} = W_{\text{atom}_1}{\mathbf{h}_i} + W_{\text{in-deg}_1}\text{deg}_i^- + W_{\text{out-deg}_1}\text{deg}_i^+.
\end{equation}

\vspace{-0.2cm}
\noindent \textbf{2. Bond Encoding} \quad
We incorporate pair-wise atom information into the edge encoding with message-passing mechanism. For every edge $\mathbf{e}_{ij}$, we use a \textit{permutation-invariant} function to generate the embedded edge representations, ensuring consistency in the learned representation regardless of the order of the atoms,
\begin{equation}
    \mathbf{z}_{\mathbf{e}_{ij}} = W_{\text{comb}_1}([W_{\text{atom}_2}\mathbf{h}_{i} + W_{\text{edge}_1}\mathbf{e}_{ij} + W_{\text{atom}_2}\mathbf{h}_{j}]) + b_{\text{comb}_1}.
\end{equation}
In addition, to ensure the symmetry of edge encoding, we calculate it as $\mathbf{z}_{\mathbf{e}_{ij}} = {(\mathbf{z}_{\mathbf{e}_{ij}} + \mathbf{z}_{\mathbf{e}_{ji}})}/{2}$.

\begin{figure*}[t]
\centering
{
\includegraphics[width=1.\textwidth]{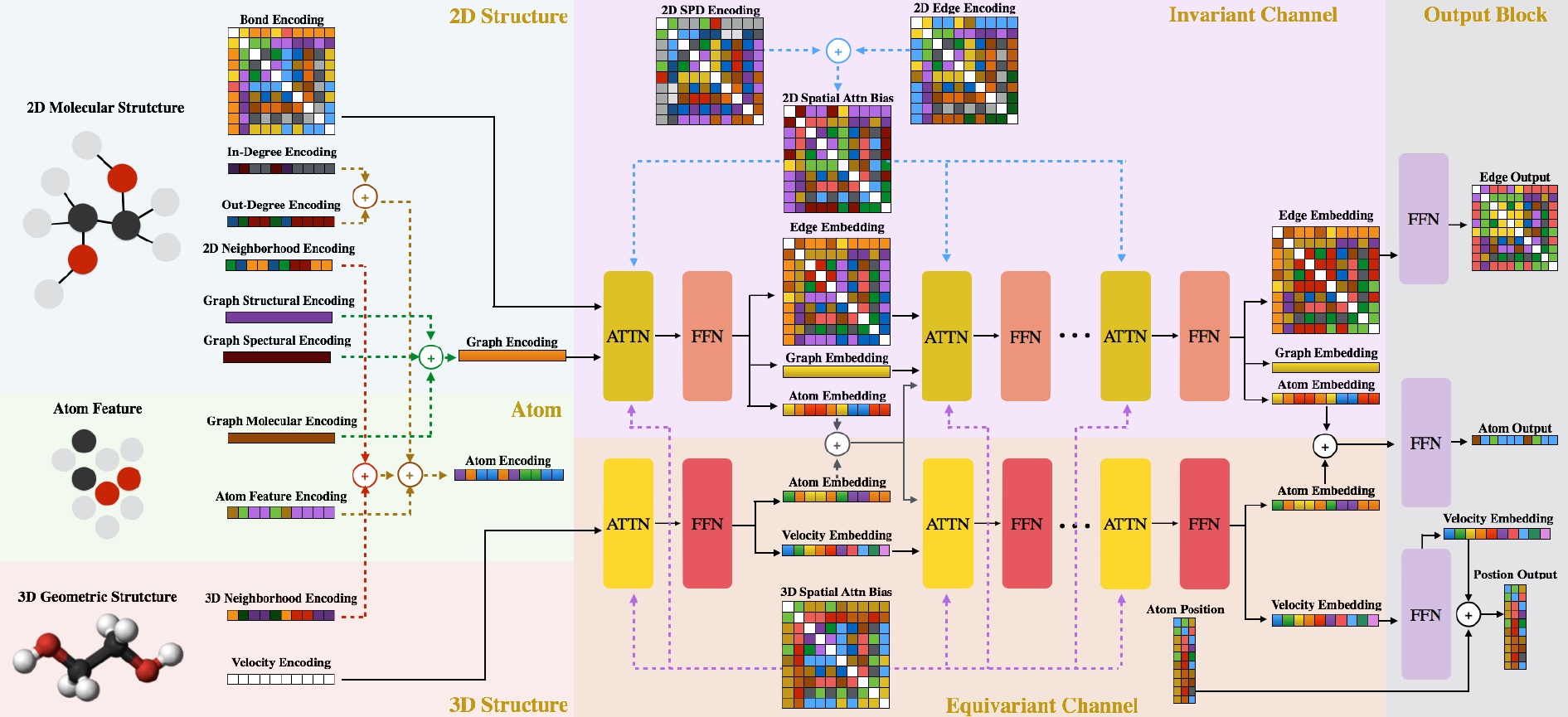}}
{%
\vspace{-0.5cm}
  \caption{The figure showcases our MUformer for processing 2D and 3D molecular data. Within the Transformer backbone, two channels exist: purple for 2D data and brown for 3D data. The blue part encodes 2D molecular structures, while the green part handles atom-level information and the red part processes 3D geometric structures. With missing 2D or 3D structures, the model activates either the invariant (purple) or equivariant (brown) channel. The invariant channel predicts atom and edge features, while the equivariant channel offers geometric transformation robustness and predicts atom features and positions. When both channels are operational, the model maintains robustness to geometric transformations and predicts a complete molecule, and final atom features are derived by merging outputs from both channels and feeding the combined data through an output network.}%
  \label{fig:mudiff}
}
\vspace{-0.7cm}
\end{figure*}

\vspace{-0.2cm}
\noindent \textbf{3. Graph Encoding} \quad
We encode graph-level structural $\mathbf{h}_{\text{struct}}$, spectral $\mathbf{h}_{\text{spect}}$ and molecular information $\mathbf{h}_{\text{mol}}$ to a molecule ${\mathbf{M}}$. As suggested in \cite{beaini2021directional, vignac2022digress}, we add cycle counts for $\mathbf{h}_{\text{struct}}$, the number of cycles of up to size 6 and the number of connected components; and we add eigenvalue features to $\mathbf{h}_{\text{spect}}$, including the multiplicity of eigenvalue 0, as well as the first 5 nonzero eigenvalues. For $\mathbf{h}_{\text{mol}}$, we include the current valency of each atom and the current molecular weight of the entire molecule as features. For every molecule $\textbf{M}$, the graph-level representation is given by,
\begin{equation}
    \mathbf{Z_M} = W_{\text{graph}}([\mathbf{h}_{\text{struct}}, \mathbf{h}_{\text{spect}}, \mathbf{h}_{\text{mol}}]) + b_{\text{graph}}.
\end{equation}

\vspace{-0.2cm}
\noindent \textbf{4. 2D Neighborhood Encoding} \quad
To get local neighborhood information, we use message passing to aggregate information from the immediate neighbors of each atom in the 2D graph $\mathbf{M}^{\text{2D}}$. Specifically, for $\mathbf{h}_i \in \mathbf{H}$ and $\mathbf{e}_{ij} \in E$, the aggregated representation of atom $i$ is calculated by
\begin{equation}
\begin{aligned}
    &\mathbf{m}_{j\to i} = \left( W_{\text{atom}_3}\mathbf{h}_j \right) \odot \left( W_{\text{edge}_2}\mathbf{e}_{ij} \right)  , \ \mathbf{z}^{\text{2D}}_{\mathbf{h}_i} = W_{\text{atom}_3}\mathbf{h}_i + \frac{1}{|N(i)|}\sum_{j\in N(i)} \mathbf{m}_{j\to i} ,
\end{aligned}
\end{equation}
where $N(i)$ denotes the neighbors of atom. To formulate the messages $\mathbf{m}$, we use the Hadamard product, $\odot$, between the atom embedding and edge embeddings. 

\vspace{-0.1cm}
\noindent \textbf{5. 3D Neighborhood Encoding} \quad
We use message passing to collect the atom information in the vicinity of the given atom to get the 3D neighborhood encoding $\mathbf{M}^{\text{3D}}$. For atom $i$, it follows
\begin{equation}
\begin{aligned}
\label{eq:3d.neigh.encoding}
    & {d}_{ij} = \|{x}_i - {x}_j\|_2, \ \mathbf{e}_{ij} = \text{SiLU} \left(W_{\text{edge}_3}\left(f_{\text{RBF}_1}\left( d_{ij} \right)\right) \right) \odot f_{\cos}(d_{ij}) \\
    &\mathbf{m}_{j\to i}  = (W_{\text{atom}_4}\mathbf{h}_j) \odot \mathbf{e}_{ij} , \ \mathbf{z}^{\text{3D}}_{\mathbf{h}_i} = W_{\text{atom}_4}\mathbf{h}_i + \frac{1}{|N(i)|}\sum_{j\in N(i)} \mathbf{m}_{j\to i}.
\end{aligned}
\end{equation}
where $d_{ij}$ is the Euclidean distance between atoms $i,j$ and $f_{\text{RBF}_1}(\cdot)$ is the $\exp$ radial basis function. To calculate messages $\mathbf{m}$, we use the Hadamard product, $\odot$, between the atom and edge embeddings.

We use the cosine cutoff $f_{\cos}(d_{ij})$ to determine which atoms in the Euclidean space given by $\mathbf{M}^{\text{3D}}$ should be considered as part of the neighborhood of atom $i$. This provides a smooth way to incorporate spatial locality in the message-passing process. Only atoms $j$ for which $f_{\cos}(d_{ij}) > 0$ are included in the message-passing process, resulting in a new node representation $\mathbf{z}^{\text{3D}}_{\mathbf{h}_i} \in \mathbb{R}^{fh_3}$ for atom $i$. 

\vspace{-0.1cm}
\noindent \textbf{Remark} \quad We can integrate 2D graph information by assigning $f_{\cos}(d_{ij})=1$ when an edge is present between atoms $i$ and $j$ in the 2D molecular structure $\mathbf{M}^{\text{2D}}$. By doing so, we effectively incorporate locality information from both the 2D molecular structure and the 3D geometric structure, providing a more comprehensive representation of the molecule.

\vspace{-0.1cm}
\noindent \textbf{6. Combine Encoding} \quad
In the final step of encoding, we compute the atomic embedding $\mathbf{Z_H}$ by concatenating the atom encoding, 2D neighborhood encoding, and 3D neighborhood encoding,
\begin{equation}
    \mathbf{Z_H} = W_{\text{comb}_2}([\mathbf{Z^{\text{1D}}_H}, \mathbf{Z^{\text{2D}}_H}, \mathbf{Z^{\text{3D}}_H}]) + b_{\text{comb}_2}.
\end{equation}
Additionally, the bond encoding $\mathbf{Z_E}$ and graph encoding $\mathbf{Z_M}$ are utilized to calculate attentions (visualized in Fig~\ref{fig:mudiff}), with details discussed in Sec~\ref{sec:transformer_channels} and App~\ref{app:transformer.arch}.

\vspace{-0.4cm}
\subsection{Attention Biases}
\vspace{-0.3cm}
\label{sec:attention_biases}
The MUformer incorporates 4 attention biases, which serve to encode spatial relationships in both 2D molecular structure and 3D geometric arrangement. These biases are integrated into the attention mechanism, enhancing the model's ability to process and understand molecular representations. A more complete introduction of MUformer attention biases is discussed in App~\ref{app:muformer.attention.bias}. And a detailed discussion of the importance and advantages of employing these attention biases for computing attentions in the MUformer can be found in App~\ref{app:discussion.attention.bias}.

\vspace{-0.1cm}
\noindent \textbf{1. 2D Spatial Attention Bias} \quad
To capture the structural relationships between atoms in the 2D molecular graph, we use the shortest path distance (SPD) encoding \cite{ying2021transformers}. The SPD encoding, denoted as $\Phi^{\text{2D}}_{\text{SPD}}(i,j): V\times V \to \mathbb{R}$, calculates the distance between atoms $i,j$ in $\mathbf{M}^{\text{2D}}$. Moreover, we incorporate edge information along the shortest path between atoms $i,j$ to reflect edge characteristics, 
\begin{equation}
\Phi^{\text{2D}}_{\text{ENC}_{ij}} = \frac{1}{N}\sum_{n=1}^N\mathbf{e}_n(w_n)^T \in \mathbb{R}.
\end{equation} 
The combined bias, $\Phi^{\text{2D}}_\mathbf{E}$, is calculated as the sum of $\Phi^{\text{2D}}_{\text{SPD}}$ and $\Phi^{\text{2D}}_{\text{ENC}}$, both in $\mathbb{R}^{n\times n\times 1}$. The 2D spatial attention bias enables the model to capture the intricate relationships between atoms and their surroundings, ultimately improving its performance.

\vspace{-0.1cm}
\noindent \textbf{2. 3D Spatial Attention Bias} \quad
To encode 3D spatial relationships between atom pairs in $\mathbf{M}^{\text{3D}}$, we use Euclidean distance and an exponential radial basis function, $f_{\text{RBF}2}(\cdot)$ \cite{schutt2017schnet}. This 3D spatial attention bias, $\Phi^{\text{3D}}_{\mathbf{E}}$, enables the model to account for the geometric arrangement of atoms in the molecule, which is crucial for understanding 3D structure and molecular properties.
\begin{equation}
\begin{aligned}
    & {d}_{ij} = \|{x}_i - {x}_j\|_2 , \ \Phi^{\text{3D}}_{\mathbf{e}_{ij}} = W_{\text{3D}_2}\left(\text{SiLU}(W_{\text{3D}_1}\left(f_{\text{RBF}_2}\left( d_{ij} \right)\right))\right).
\end{aligned}
\end{equation}
 By incorporating this 3D spatial attention bias, the MUformer is able to better capture complex 3D spatial relationships between atoms, leading to improved performance in tasks involving 3D molecular structures.

\vspace{-0.2cm}
\noindent \textbf{Edge Feature \& Graph Feature} \quad
The embeddings obtained from the 2D structure, $\mathbf{Z_E}$, and the molecular graph information, $\mathbf{Z_M}$, can be further projected and employed as additional attention biases, enhancing the model's understanding of the molecular structure and relationships.

\vspace{-0.4cm}
\subsection{Transformer Channels}
\vspace{-0.2cm}
\label{sec:transformer_channels}
The MUformer employs two channels, the \textit{invariant} and the \textit{equivariant channel}, to process molecular data, and an output block to combine the processed information from the two channels
 (see Fig~\ref{fig:mudiff}). A more complete introduction of MUformer transformer channels is discussed in App~\ref{app:transformer.arch}.

\vspace{-0.1cm}
\noindent \textbf{Invariant Channel} \quad
The invariant channel in MUformer captures intrinsic properties of the input 2D molecule graph, $\mathbf{M}^{\text{2D}}$, but also utilizes 2D and 3D spatial biases as attention biases when 3D geometric structure is provided. Its main goal is to predict invariant features, including atom and edge-level features, by leveraging the underlying graph structure. Unlike in \cite{ying2021transformers}, the invariant channel in MUformer is used to predict atom and edge features, generating embeddings for invariant features $\hat{\mathbf{Z}}_{\mathbf{H}}$ and $\hat{\mathbf{Z}}_{\mathbf{E}}$, respectively (see purple part in Fig~\ref{fig:mudiff}).

\vspace{-0.1cm}
\noindent \textbf{Equivariant Channel} \quad
The equivariant channel in MUformer extracts features from the input 3D molecule graph $\mathbf{M}^{\text{3D}}$ that can vary under geometric transformations. Unlike \cite{tholke2022torchmd}, which only focuses on the 3D molecular structure, our equivariant channel can utilize both 2D molecular structure and 3D geometric structure in Euclidean space (if the 2D molecular structure is provided). This enables the channel to predict atom features $\hat{\mathbf{Z}}_{\mathbf{H}}$ and their coordinates $\hat{\mathbf{Z}}_{\mathbf{X}}$ that are insensitive to geometric transformations (see the brown part in Fig~\ref{fig:mudiff}). The unique capability of our equivariant channel enables it to capture richer structural information from both 2D and 3D molecular representations.

\vspace{-0.1cm}
\noindent \textbf{Interaction Embedding} \quad
The atom representations $\hat{\mathbf{Z}}_\mathbf{H}$ from two transformer channels are combined, and fed in to the next layer of the transformer channels or used as input for the final predictions,
\begin{equation}
    \mathbf{\hat{Z}_H} = {W}_{\text{comb}_3}[\mathbf{Z}_{\mathbf{H}}^{\text{inv}}, \mathbf{Z}_{\mathbf{H}}^{\text{eqv}}] + {b}_{\text{comb}_3}.
\end{equation}

\vspace{-0.3cm}
\noindent \textbf{Output Block} \quad
The output block generates the final output utilizing the embeddings from invariant and equivariant channels. Specifically, it takes in the atom $\hat{\mathbf{Z}}_{\mathbf{H}}$ and edge embeddings $\hat{\mathbf{Z}}_{\mathbf{E}}$, along with the velocity embedding $\hat{{\vv}}$. Through feature extractions, the output block makes predictions,
\begin{equation}
\begin{aligned}
    & \mathbf{H}_{\text{out}} = W_{{\mathbf{X}}_{\text{out}_2}}\left(\text{SiLU}(W_{{\mathbf{X}}_{\text{out}_1}}\hat{\mathbf{Z}}_{\mathbf{H}})\right) \in \mathbb{R}^{n\times d_{\text{out}}}, \mathbf{X}_{\text{out}} = \mathbf{X} + W_{{{\vv}}_{\text{out}_2}}\left(\text{SiLU}(W_{{{\vv}}_{\text{out}_1}}\hat{{\vv}})\right) \in \mathbb{R}^{n\times 3} \\
    & \mathbf{E}_{\text{out}} = W_{{\mathbf{E}}_{\text{out}_2}}\left(\text{SiLU}(W_{{\mathbf{E}}_{\text{out}_1}}\hat{\mathbf{Z}}_{\mathbf{E}})\right) \in \mathbb{R}^{n\times n \times b_{\text{out}}}.
\end{aligned} 
\end{equation}

\vspace{-0.8cm}
\section{Related Work}
\label{sec:priorwork}
\vspace{-0.2cm}
\subsection{Graph Models and Equivariant Models}
\vspace{-0.2cm}
Graph models, \ie{} graph neural networks and graph transformers, and equivariant models have emerged as crucial tools to perform molecule-relevant tasks. Graph models are a class of neural networks specifically designed for graph-structured data \cite{kipf2016classification,hamilton2017inductive,luan2019break, luan2021heterophily, luan2022revisiting, hua2022high, luan2023graph}, such as molecular structures, and have demonstrated effectiveness in capturing the complex relationships between atoms and bonds within a molecule \cite{chen2020can, dwivedi2020generalization, ying2021transformers, tholke2022torchmd}. Equivariance is a property of a function $f$ \wrt{} a group, such that the function preserves its structure under group transformations, mathematically expressed as $f(g x) = g f(x)$ for any $g \in G$ and input $x$ \cite{satorras2021n}. Equivariant graph models constitute a subcategory of GNNs that exhibit equivariance with respect to a group of symmetries, enabling them to learn representations invariant to specific transformations such as translation, rotation, and reflection. This makes them well-suited for tasks involving geometric structure \cite{satorras2021n, satorras2021n2, tholke2022torchmd}. The integration of graph models and equivariance in graph-structured data has led to significant advancements in the design and discovery of molecules.

\vspace{-0.4cm}
\subsection{Diffusion Models}
\vspace{-0.2cm}
Diffusion models have gained significant traction as powerful generative models for drug discovery applications \cite{ho2020denoising, kingma2021variational, austin2021structured}. For instance, \cite{xu2022geodiff} employs diffusion models to generate molecules with the lowest conformation energy. In another work, \cite{vignac2022digress} presents a diffusion model that utilizes a graph transformer to denoise the diffusion process, operating jointly on atom features and molecular structures.
Moreover, recent studies have explored the integration of equivariant graph models within diffusion models for molecule generation. For example, \cite{hoogeboom2022equivariant} introduces a diffusion model with an equivariant GNN, enabling the model to work jointly on atom features and coordinates. Overall, incorporating equivariant graph models into diffusion models for molecule generation has demonstrated promising results in recent research. 

\vspace{-0.4cm}
\subsection{Joint Generative Models}
\vspace{-0.2cm}
\cite{zhang2023equivariant} employs an autoregressive flow as the backbone model to generate atom types, bond types, and 3D coordinates sequentially. For the atomic coordinates, \cite{zhang2023equivariant} constructs a local spherical coordinate system based on local reference atoms and predicts the relative coordinates. \cite{peng2023moldiff} tackles the atom-bond inconsistency problem in 3D molecule generation by employing a diffusion model that generates atoms and bonds simultaneously while maintaining their consistency. \cite{peng2023moldiff}  incorporates a noise schedule to gradually add noise to the atom positions and types, as well as bond types with a guidance, to perturb them towards the correct values. Moreover, \cite{zhang2022molecule} works on the structure-based drug design that generate both 2D and 3D molecular graphs to enhance the overall representation.

\vspace{-0.5cm}
\section{Experiments}
\vspace{-0.2cm}
\label{sec:experiments}
To evaluate our MUDiff framework, we conduct experiments on the QM9 dataset \cite{ramakrishnan2014quantum}, which contains 130k small molecules with up to 9 heavy atoms (29 atoms including hydrogens) and their associated molecular properties and structures.
We use the train/val/test splits from \cite{anderson2019cormorant}, consisting of 100K/18K/13K samples respectively, for evaluation. This protocol follows the method used in previous works such as \cite{satorras2021n, hoogeboom2022equivariant}. 

\vspace{-0.4cm}
\subsection{Molecule Generation with Limited 3D Data}
\vspace{-0.3cm}
\label{sec:molecule.generation.limit.3d}
In this section, we introduce a new molecule generation task that incorporates limited 3D data, as many real-world datasets lack complete 3D structures.
\begin{wraptable}{r}{0.46\textwidth}
\centering
\vspace{-0.5cm}
  \caption{Negative log-likelihood, atom stability, and molecule stability are evaluated with standard deviation across 3 runs on QM9, using 10K samples from the model. 30K+70K means model trained with limited 3D data.}
  \vspace{-0.2cm}
  \resizebox{6.0cm}{!}{
    \footnotesize
    \begin{tabular}{lccc}
    \hline
    \hline
    \textbf{Method} & NLL & Atom Stable(\%) & Mol Stable(\%)\\
    \hline
    \textbf{EDM} & -$110.7 \pm 1.5$ & $ {98.7} \pm 0.1 $ & $82.0 \pm 0.4 $\\
    \textbf{DiGress} & - & $98.1 \pm 0.3 $  & $79.8 \pm 5.6 $\\
    \textbf{MUDiff} & ${-135.5} \pm 2.1 $  &   $ {98.8} \pm 0.2 $    &  $ {89.9} \pm 1.1 $\\
    \textbf{MUDiff} (30K+70K) & ${-120.6} \pm 3.4 $  &   $ {98.2} \pm 0.7 $    &  $ {84.5} \pm 2.5 $\\
    \hline
    \hline
    \end{tabular}%
    }
  \label{tab:limited.3D.data}%
  \vspace{-0.6cm}
\end{wraptable} 
To accomplish this, we randomly split the 100K training molecules into two sets: 30K with both 2D and 3D structures and 70K with only 2D structures. We train the model on the 30K samples using both the invariant and equivariant channels and validate on 18K samples until NLL converges.
We then fine-tune the trained model on the remaining 70K molecules with only 2D structures and validate/test on 18K/13K samples.

\vspace{-0.2cm}
\noindent \textbf{Remark} \quad
Notably, this training framework with limited 3D data is only possible with MUDiff for now, because of the flexible two-channel design.

\vspace{-0.2cm}
\noindent \textbf{Results} \quad
The results of the molecule generation task with limited 3D data are summarized in Table~\ref{tab:limited.3D.data}. MUDiff achieved competitive results in generating stable molecules, even with limited 3D information in the training set, compared to the baselines. These results suggest that MUDiff has the ability to leverage sufficient 2D structures to infer 3D geometry. This finding may motivate further research on the co-generation of 2D and 3D structures for molecules.

\vspace{-0.4cm}
\subsection{Molecule Generation}
\vspace{-0.3cm}
\label{sec:molecule.generation}
\begin{wraptable}{r}{0.46\textwidth}
\centering
\vspace{-0.5cm}
  \caption{Negative log-likelihood, atom stability, and molecule stability are evaluated with standard deviation across 3 runs on QM9, using 10K samples (with hydrogen) from the model. The results surpass those of previous models, as reported in \cite{hoogeboom2022equivariant, vignac2022digress}. }
  \vspace{-0.2cm}
  \resizebox{6.0cm}{!}{
    \footnotesize
    \begin{tabular}{lccc}
    \hline
    \hline
    \textbf{Method} & NLL & Atom Stable(\%) & Mol Stable(\%)\\
    \hline
    \textbf{Data} & - & $99.0$   & $95.2 $\\
    \hline
    \textbf{ENF} & -$59.7$ & $85.0 $   & $4.9 $\\
    \textbf{GSchnet} & - & $95.7$  & $68.1 $\\
    \textbf{GDM} & -$92.5$ & $97.6 $ & $71.6 $\\
    \textbf{EDM} & -$110.7 \pm 1.5$ & $ {98.7} \pm 0.1 $ & $82.0 \pm 0.4 $\\
    \textbf{DiGress} & - & $98.1 \pm 0.3 $  & $79.8 \pm 5.6 $\\
    \textbf{MDM} & - & $98.6 $  & $\textbf{91.9} $\\
    \textbf{GeoLDM} & - & $\textbf{98.9} \pm 0.1 $  & $89.4 \pm 0.5 $\\
    \textbf{MUDiff} (ours) & $\textbf{-135.5} \pm 2.1 $  &   $ \textbf{98.8} \pm 0.2 $    &  $ \textbf{89.9} \pm 1.1 $\\
    \hline
    \hline
    \end{tabular}%
    }
  \label{tab:cond.generation.stable}%
  \vspace{-0.5cm}
\end{wraptable} 
We compare the performance of our MUDiff 
model with popular generative models, including GraphVAE \cite{kipf2016variational}, GSchenet \cite{gebauer2019symmetry}, Set2GraphVAE \cite{vignac2021top}, ENF \cite{satorras2021n}, GDM \cite{hoogeboom2022equivariant}, EDM \cite{hoogeboom2022equivariant}, DiGress \cite{vignac2022digress}, MDM \cite{huang2023mdm}, and GeoLDM \cite{xu2023geometric}. 
The results of the baseline models can be found in the studies by \cite{hoogeboom2022equivariant} and \cite{vignac2022digress}.

As outlined in \cite{satorras2021n}, we evaluate the atom and molecule stability of the generated compounds by measuring the proportion of atoms that have the correct valency for atom stability, and the proportion of generated molecules in which all atoms are stable for molecule stability. Additionally, we also measure the validity and uniqueness using the RDKit tool, as used in \cite{hoogeboom2022equivariant}. 

\vspace{-0.1cm}
\noindent \textbf{Remark} \quad
We would like to emphasize that the dataset statistics are not ideal, with atom stability at 99\%, molecule stability at 95.2\%, and molecule validity at 99.3\% in the original data. 
These statistics are not perfect, pointing to potential imperfections in the dataset. The imperfections of the dataset have also been acknowledged in \cite{satorras2021n, hoogeboom2022equivariant, vignac2022digress}. Additionally, the metrics used in this study have their own limitations, which are discussed in App~\ref{app:molecule.generation}.

\begin{wraptable}{r}{0.46\textwidth}
  \centering
  \vspace{-1cm}
  \caption{Validity and uniqueness of over 10K molecules are shown with standard deviation across 3 runs, surpassing the results of previous models according to studies by \cite{hoogeboom2022equivariant, vignac2022digress}. }
  \vspace{-0.2cm}
  \resizebox{5.7cm}{!}{
  \footnotesize
    \begin{tabular}{lccc}
    \hline
    \hline
    \textbf{Method} & w/ Hydrogen & Valid (\%) & Unique (\%) \\
    \hline
    \textbf{Data}  & & $99.3$  & $100.0$ \\
    \hline
    \textbf{GraphVAE} & & $55.7$  & $42.3$ \\
    \textbf{Set2GraphVAE} &  & $59.9 \pm 1.7$  & $56.2 \pm 1.4$ \\
    \textbf{EDM}   & & $97.5 \pm 0.2$  & $94.3 \pm 0.2$ \\
    \textbf{DiGress} &  & $ \textbf{99.0} \pm 0.1$    & $96.2 \pm 0.1$ \\
    \textbf{MUDiff} (ours) &   &   $ \textbf{98.9} \pm 0.4$  &  $ \textbf{99.3} \pm 0.3$ \\
    \hline
    \textbf{Data}  & \checkmark & $97.8$  & $100.0$ \\
    \hline
    \textbf{ENF}   & \checkmark & $40.2$  & $39.4$ \\
    \textbf{GSchnet} & \checkmark & $85.5$  & $80.3$ \\
    \textbf{GDM}   & \checkmark & $90.4$  & $89.5$ \\
    \textbf{EDM}   & \checkmark & $91.9 \pm 0.5$  & $90.7 \pm 0.6$ \\
    \textbf{DiGress} & \checkmark & $\textbf{95.4} \pm 1.1$  & $97.6 \pm 0.4$ \\
    \textbf{GeoLDM} & \checkmark & ${93.8} \pm 0.4$  & $92.7 \pm 0.5$ \\
    \textbf{MUDiff} (ours) & \checkmark &  $\textbf{95.3} \pm 1.5$    &  $\textbf{99.1} \pm 0.5$ \\
    \hline
    \hline
    \end{tabular}%
    }
  \label{tab:cond.generation.valid}%
  \vspace{-0.5cm}
\end{wraptable} 

\vspace{-0.1cm}
\noindent \textbf{Results} \quad
Table~\ref{tab:cond.generation.stable} presents the evaluated results of atom and molecule stability for molecules generated by MUDiff and the baseline models. The reported average results and standard deviations are over 3 runs, using 10,000 samples from each model. The table shows that MUDiff can generate molecules that are significantly more stable than the baseline models in terms of negative log-likelihood and molecule stability and matches the performance of SOTA model \wrt{} atom stability.

Table~\ref{tab:cond.generation.valid} presents the results of the validity and uniqueness of the generated samples. It should be noted that, following the guidelines outlined in \cite{vignac2021top}, novelty is not reported in this table. The table shows that MUDiff generates a significantly higher rate of unique molecules than the baselines and matches the rate of valid molecules of SOTA models.

\vspace{-0.4cm}
\subsection{Conditional Generation}
\vspace{-0.2cm}
We follow the experimental setting in \cite{hoogeboom2022equivariant} to train the conditional MUDiff on the QM9 dataset, conditioning the generation on properties $\alpha$, $\epsilon_{\text{homo}}$, $\epsilon_{\text{lumo}}$, $\Delta\epsilon$, $\mu$, and $C_v$, respectively.

\vspace{-0.4cm}
\subsection{Ablation Study}
\vspace{-0.2cm}
To investigate how different components affect the performance of MUDiff, we conduct an ablation study in App~\ref{app:ablation.study}. Further details about the experimental settings can be found in App~\ref{app:ablation}.

\vspace{-0.4cm}
\subsection{Property Prediction}
\vspace{-0.2cm}
Additionally, we conduct a comprehensive comparison of our MUformer with several baselines on the QM9 dataset for property prediction, including SchNet \cite{schutt2017schnet}, EGNN \cite{satorras2021n}, PhysNet \cite{unke2019physnet}, DimeNet \cite{beani2021directional}, Cormorant \cite{anderson2019cormorant}, PaiNN \cite{schutt2021equivariant}, and ET \cite{tholke2022torchmd}. The dataset consists of molecules with various properties, and we estimate 12 chemical properties per molecule following \cite{satorras2021n}. The comparison results in terms of mean absolute error are presented in App~\ref{app:qm9.property}.

\vspace{-0.6cm}
\section{Conclusion}
\vspace{-0.2cm}
In this work, we introduced MUDiff, a transformer-based framework for learning and generating a complete molecule representation using a novel architecture, MUformer. Our contributions include proposing a new molecule generation method, named MUDiff, which successfully generates more stable and valid molecules, demonstrating the potential for further research and applications. We also explored the interplay between 2D and 3D structure generation in App~\ref{app:discussion.2d3d.relation}, which reveals the benefits of jointly generating both structures to enhance the overall performance. Additionally, we discussed the scalability issues in App~\ref{app:discussion} and provided insights for future work to improve the efficiency and accuracy of MUDiff. By addressing these challenges, we aim to support progress in machine learning for molecules and facilitate advancements in areas such as drug discovery and material design.

\clearpage
\section{Acknowledgements}
This work is supported by the Natural Sciences and Engineering Research Council of Canada (NSERC) Grant, Canadian Institute for Advanced Research (CIFAR) Grant, and Fonds d’accélération des collaborations en santé (FACS-Acuity) supported by Ministre de lconomie et de lInnovation Canada. Minkai Xu thanks the generous support of Sequoia Capital Stanford Graduate Fellowship. 
\bibliographystyle{unsrtnat}
\bibliography{reference}

\clearpage
\appendix
\section{MUformer}
\label{app:muformer}

\subsection{MUformer Encodings}
\label{app:muformer.encoding}
The MUformer consists of 6 encoding functions, with 3 being message-passing based, designed to incorporate atomic, positional, and structural information into a concise and expressive representation, particularly suited for handling graph-structured inputs.

\noindent \textbf{1. Atom Encoding} \quad
Incorporating centrality information into the atom representations is crucial as it helps to highlight the importance of individual atoms in the molecular structure. The authors in \cite{ying2021transformers} propose a method of utilizing in-degree $\text{deg}^-$ and out-degree $\text{deg}^+$ obtained from 2D molecular graphs $\mathbf{M}^{\text{2D}}$ to incorporate centrality information into the atom-wise encoding process. This allows for a detailed and accurate representation of the molecular structure, taking into account the relative importance of each atom. After the centrality encoding, the new representation $\mathbf{z}^{\text{1D}}_{\mathbf{h}_i}$ for node $i$ is,
\begin{equation}
    \mathbf{z}^{\text{1D}}_{\mathbf{h}_i} = W_{\text{atom}_1}{\mathbf{h}_i} + W_{\text{in-deg}_1}\text{deg}_i^- + W_{\text{out-deg}_1}\text{deg}_i^+,
\end{equation}
where $W_{\text{atom}_1} \in \mathbb{R}^{d \times fh_1}, W_{\text{in-deg}_1}  \in \mathbb{R}^{1 \times fh_1}, W_{\text{out-deg}_1}  \in \mathbb{R}^{1 \times fh_1}$ are designated learnable parameters for the atom features, in-degree $\text{deg}^-$, and out-degree $\text{deg}^+$. The resulting atom embedding for atom $i$, $\mathbf{z}^{\text{1D}}_{\mathbf{h}_i} \in \mathbb{R}^{fh_1}$, includes the degree centrality information.

\noindent \textbf{2. Bond Encoding} \quad
To obtain a richer edge representation, we incorporate the pair-wise atom information into the edge encoding with message-passing information. For each edge $\mathbf{e}_{ij} \in \mathbf{E}$, we use a \textit{permutation-invariant} function to generate the embedded edge representations, ensuring consistency in the learned representation regardless of the order of the atoms,
\begin{equation}
    \mathbf{z}_{\mathbf{e}_{ij}} = W_{\text{comb}_1}([W_{\text{atom}_2}\mathbf{h}_{i} + W_{\text{edge}_1}\mathbf{e}_{ij} + W_{\text{atom}_2}\mathbf{h}_{j}]) + b_{\text{comb}_1}.
\end{equation}
where $W_{\text{atom}_2} \in \mathbb{R}^{d \times fe_{\text{in}}}$ and $W_{\text{edge}_1} \in \mathbb{R}^{b \times fe_{\text{in}}}$ are learnable parameters that handle the atom features and edge types respectively, $W_{\text{comb}_1} \in \mathbb{R}^{fe_{\text{in}} \times fe_{\text{in}}}$ combines the representations with bias $b_{\text{comb}_1}$. In addition, in order to make the edge representation symmetric, we calculate the edge representation as $\mathbf{z}_{\mathbf{e}_{ij}} = {(\mathbf{z}_{\mathbf{e}_{ij}} + \mathbf{z}_{\mathbf{e}_{ji}})}/{2}$. The resulting edge embedding is $\mathbf{Z_E}\in \mathbb{R}^{n\times n \times fe_{\text{in}}}$.

\noindent \textbf{3. Graph Encoding} \quad
Standard GNNs have limitations in detecting simple substructures such as cycles \cite{chen2020can}, which can hinder their ability to accurately capture the properties of the data distribution. To overcome this limitation, we enhance our model by extra features as follows.

In the graph encoding process, we encode graph-level structural $\mathbf{h}_{\text{struct}}$, spectral $\mathbf{h}_{\text{spect}}$ and molecular information $\mathbf{h}_{\text{mol}}$ to a molecule ${\mathbf{M}}$. As suggested in \cite{beaini2021directional, vignac2022digress}, we add cycle counts for $\mathbf{h}_{\text{struct}}$, the number of cycles of up to size 6 and the number of connected components; and we add eigenvalue features to $\mathbf{h}_{\text{spect}}$, including the multiplicity of eigenvalue 0, as well as the first 5 nonzero eigenvalues. For $\mathbf{h}_{\text{mol}}$, we include the current valency of each atom and the current molecular weight of the entire molecule as features. For every molecule $\textbf{M}$, the graph-level representation is given by,
\begin{equation}
    \mathbf{Z_M} = W_{\text{graph}}([\mathbf{h}_{\text{struct}}, \mathbf{h}_{\text{spect}}, \mathbf{h}_{\text{mol}}]) + b_{\text{graph}},
\end{equation}
where $W_{\text{graph}} \in \mathbb{R}^{13 \times f_{\text{in}}}$ combines the encoded information with bias $b_{\text{graph}}$. The resulting graph representation, $\mathbf{Z_M} \in \mathbb{R}^{1\times f_{\text{in}}}$, encapsulates all the information from the structural, spectral and molecular features.

\noindent \textbf{4. 2D Neighborhood Encoding} \quad
To get local neighborhood information, we use message passing to aggregate information from the immediate neighbors of each atom in the 2D graph $\mathbf{M}^{\text{2D}}$. Specifically, for $\mathbf{h}_i \in \mathbf{H}$ and $\mathbf{e}_{ij} \in E$, the aggregated representation of atom $i$ is calculated by
\begin{equation}
\begin{aligned}
    &\mathbf{m}_{j\to i} = \left( W_{\text{atom}_3}\mathbf{h}_j \right) \odot \left( W_{\text{edge}_2}\mathbf{e}_{ij} \right)  , \ \mathbf{z}^{\text{2D}}_{\mathbf{h}_i} = W_{\text{atom}_3}\mathbf{h}_i + \frac{1}{|N(i)|}\sum_{j\in N(i)} \mathbf{m}_{j\to i} ,
\end{aligned}
\end{equation}
where $N(i)$ denotes the neighbors of atom $i$, $W_{\text{atom}_3} \in \mathbb{R}^{d \times fh_2} $ and $W_{\text{edge}_2} \in \mathbb{R}^{b \times fh_2}$ are learnable parameters for atom features and edge types, respectively. To formulate the messages $\mathbf{m}$, we use the Hadamard product, $\odot$, between the atom embedding and edge embedding. This new representation $\mathbf{z}^{\text{2D}}_{\mathbf{h}_i}\in \mathbb{R}^{fh_2}$ takes into account not only the atom features, but also the features of its neighboring atoms and the edges connecting them.

\noindent \textbf{5. 3D Neighborhood Encoding} \quad
We use message passing to collect the atom information in the vicinity of the given atom to get the 3D neighborhood encoding $\mathbf{M}^{\text{3D}}$ as suggested by \cite{satorras2021n}. For atom $i$, its representation $\mathbf{z}^{\text{3D}}_{\mathbf{h}_i}$ is computed by following steps,
\begin{equation}
\begin{aligned}
\label{app.eq:3d.neigh.encoding}
    & {d}_{ij} = \|{x}_i - {x}_j\|_2, \ \mathbf{e}_{ij} = \text{SiLU} \left(W_{\text{edge}_3}\left(f_{\text{RBF}_1}\left( d_{ij} \right)\right) \right) \odot f_{\cos}(d_{ij}) \\
    &\mathbf{m}_{j\to i}  = (W_{\text{atom}_4}\mathbf{h}_j) \odot \mathbf{e}_{ij} , \ \mathbf{z}^{\text{3D}}_{\mathbf{h}_i} = W_{\text{atom}_4}\mathbf{h}_i + \frac{1}{|N(i)|}\sum_{j\in N(i)} \mathbf{m}_{j\to i}.
\end{aligned}
\end{equation}
where $d_{ij}$ is the distance from atom $i$ to $j$ in the Euclidean space, $f_{\text{RBF}_1}(\cdot)$ is the exponential radial basis function, $f_{\cos}(\cdot)$ is the cosine cutoff, SiLU($\cdot$) is the activation function, $\odot$ denotes the Hadamard product, $W_{\text{atom}_4} \in \mathbb{R}^{d\times fh_3}$ and $W_{\text{edge}_3} \in \mathbb{R}^{k \times fh_3 }$ are designated learnable parameters for atom features and edge features, $k$ is the number of basis kernels as mentioned in Sec~\ref{sec:basics}. To calculate messages $\mathbf{m}$, we use the Hadamard product, $\odot$, between the atom and edge embeddings.

We use the cosine cutoff $f_{\cos}(d_{ij})$ to determine which atoms in the Euclidean space given by $\mathbf{M}^{\text{3D}}$ should be considered as part of the neighborhood of atom $i$. This provides a smooth and differentiable way to incorporate spatial locality in the message-passing process, focusing on atoms that are closer in the 3D space while ignoring distant ones. This allows the model to better capture local geometric information and reduce computational complexity by not considering interactions between atoms that are too far apart, which would be less relevant for the molecular properties under investigation. Only atoms $j$ for which $f_{\cos}(d_{ij}) > 0$ are included in the message passing aggregation process, resulting in a new node representation $\mathbf{z}^{\text{3D}}_{\mathbf{h}_i} \in \mathbb{R}^{fh_3}$ for atom $i$. 

\noindent \textbf{Remark} \quad We can integrate 2D graph information by assigning $f_{\cos}(d_{ij})=1$ when an edge is present between atoms $i$ and $j$ in the 2D molecular structure $\mathbf{M}^{\text{2D}}$. This ensures that messages between these atoms are not influenced by the smooth transition. By doing so, we effectively incorporate locality information from both the 2D molecular structure and the 3D geometric structure, providing a more comprehensive representation of the molecule.

\noindent \textbf{6. Combine Encoding} \quad
In the final step of the encoding process, we compute the atomic embedding $\mathbf{Z_H}$ by concatenating the centrality embedding, 2D neighborhood embedding, and 3D neighborhood embedding. This concatenated representation is then passed through a learnable parameter, $W_{\text{comb}2} \in \mathbb{R}^{(fh_1 + fh_2 + fh_3) \times f{\text{in}}}$. The final equation for this calculation is given by
\begin{equation}
    \mathbf{Z_H} = W_{\text{comb}_2}([\mathbf{Z^{\text{1D}}_H}, \mathbf{Z^{\text{2D}}_H}, \mathbf{Z^{\text{3D}}_H}]) + b_{\text{comb}_2}.
\end{equation}
This combination of different embeddings, $\mathbf{Z_H}\in\mathbb{R}^{n\times f_{\text{in}}}$, enables the incorporation of various molecular structure features, including centrality, 2D, and 3D neighborhood information, into a unified, comprehensive representation.

Additionally, the bond encoding $\mathbf{Z_E}$ and graph encoding $\mathbf{Z_M}$ are utilized to calculate attentions, with details discussed in Sec~\ref{sec:transformer_channels} and App~\ref{app:transformer.arch}.

\subsection{Attention Biases}
\label{app:muformer.attention.bias}
The MUformer incorporates 4 attention biases, which serve to encode spatial relationships in both 2D molecular structure and 3D geometric arrangement. These biases are integrated into the attention mechanism, enhancing the model's ability to process and understand molecular representations. A detailed discussion of the importance and advantages of employing these attention biases for computing attentions in the MUformer can be found in App~\ref{app:discussion.attention.bias}.

\noindent \textbf{1. 2D Spatial Attention Bias} \quad
To encode the structural relationships between atoms in the molecule graph $\mathbf{M}^{\text{2D}}$, we usethe shortest path distance (SPD) encoding  \cite{ying2021transformers}, denoted as $\Phi^{\text{2D}}_{\text{SPD}}(i,j): V\times V \to \mathbb{R}$, which calculates the distance between atoms $i$ and $j$ in $\mathbf{M}^{\text{2D}}$, providing valuable information about the structural relationships between atoms in the 2D molecular graph.

Additionally, following the approach of \cite{ying2021transformers}, we incorporate edge-type information along the shortest path between atoms $i$ and $j$ to reflect edge characteristics. This inclusion of edge-type information further enriches the model's understanding of the 2D molecular structure. To achieve this, we determine the shortest path $SP_{ij} = (\mathbf{e}_1, \mathbf{e}_2, ..., \mathbf{e}_N)$, where $N$ is the longest shortest path distance for all pairs of atoms $i$ and $j$. The edge encoding is computed using the following equation,
\begin{equation}
\Phi^{\text{2D}}_{\text{ENC}_{ij}} = \frac{1}{N}\sum_{n=1}^N\mathbf{e}_n(w_n)^T \in \mathbb{R} 
\end{equation}
where $w_n \in \mathbb{R}^{b \times 1}$ is a learnable vector with the same dimension as the edge feature. Both 2D spatial biases, $\Phi^{\text{2D}}_{\text{SPD}}$ and $\Phi^{\text{2D}}_{\text{ENC}}$, are in $\mathbb{R}^{n\times n\times 1}$, and the combined bias is calculated as $\Phi^{\text{2D}}_\mathbf{E} = \Phi^{\text{2D}}_{\text{SPD}} + \Phi^{\text{2D}}_{\text{ENC}}$. This 2D spatial attention bias enables the model to better capture the intricate relationships between atoms and their surroundings in the 2D molecular graph, ultimately improving its performance.

\noindent \textbf{2. 3D Spatial Attention Bias} \quad
The 3D spatial relationships between atom pairs in $\mathbf{M}^{\text{3D}}$ can be effectively encoded using the Euclidean distance and an exponential radial basis function, $f_{\text{RBF}2}(\cdot)$ \cite{schutt2017schnet}. By incorporating this 3D spatial attention bias, the model is able to account for the geometric arrangement of atoms in the molecule, which is essential for understanding the 3D structure and its impact on molecular properties. The 3D spatial bias, $\Phi^{\text{3D}}_{\mathbf{E}}$, is calculated using the following equation
\begin{equation}
\begin{aligned}
    & {d}_{ij} = \|{x}_i - {x}_j\|_2 , \ \Phi^{\text{3D}}_{\mathbf{e}_{ij}} = W_{\text{3D}_2}\left(\text{SiLU}(W_{\text{3D}_1}\left(f_{\text{RBF}_2}\left( d_{ij} \right)\right))\right),
\end{aligned}
\end{equation}
where $W_{\text{3D}_1}\in \mathbb{R}^{k \times k}, W_{\text{3D}_2}\in \mathbb{R}^{k \times 1}$ are learnable parameters, $k$ is the number of basis kernels, and the resulting 3D spatial bias, $\Phi^{\text{3D}}_{\mathbf{E}}$, is in $\mathbb{R}^{n\times n \times 1}$. By including this 3D spatial attention bias in the model, the MUformer is able to better capture the complex 3D spatial relationships between atoms, leading to improved performance in tasks involving 3D molecular structures.

\noindent \textbf{Edge Feature \& Graph Feature} \quad
The embeddings obtained from the 2D structure, $\mathbf{Z_E}$, and the molecular graph information, $\mathbf{Z_M}$, can be further projected and employed as additional attention biases, enhancing the model's understanding of the molecular structure and relationships. This allows the MUformer to better capture the complexities of the molecular system and improve its performance in various tasks. A detailed explanation of this process can be found in the subsequent section.

\subsection{Transformer Channels}
\label{app:transformer.arch}
Our MUformer architecture draws inspiration from Transformer-M \cite{luo2022one}, which employs two separate channels to process 2D and 3D molecular data, respectively. However, our model takes a different approach to processing the invariant and equivariant features of molecular data. While Transformer-M is limited to predicting atom features only, and is invariant to geometric transformations, our model can predict atom features, molecule structures, and atom positions, and is equivariant to geometric transformations. Our model can be used under different conditions of the input data. When the input data only contains 2D molecular information $\mathbf{M}^{\text{2D}}=(\mathbf{H}, \mathbf{E})$ and the geometric structure is missing, only the invariant channel is applied, and the model predicts invariant features including atom features $\mathbf{H}_{\text{out}}$ and molecular structure $\mathbf{E}_{\text{out}}$. Similarly, when the input data only contains 3D molecular information $\mathbf{M}^{\text{3D}}=(\mathbf{H}, \mathbf{X})$ and the molecular structure is missing, only the equivariant channel is used and the model becomes insensitive to geometric information, predicting atom features $\mathbf{H}_{\text{out}}$ and coordinates $\mathbf{X}_{\text{out}}$. Finally, when both 2D and 3D molecular data are provided as input, both the invariant and equivariant channels are activated. The model is equivariant to geometric transformations, predicting the complete molecule including atom features $\mathbf{H}_{\text{out}}$, molecular structure $\mathbf{E}_{\text{out}}$, and geometric structure $\mathbf{X}_{\text{out}}$.

The MUformer architecture utilizes two channels, the \textit{invariant channel} and the \textit{equivariant channel}, to learn the 2D molecular structure and 3D geometric structure, respectively. 
We simplify the notation by omitting the indices of the attention head $h$ and layer $l$.

\paragraph{Invariant Channel}
The invariant channel is an improved version of the one presented in \cite{ying2021transformers}, which is specifically designed to extract the inherent characteristics of the input molecule graph $\mathbf{M}^{\text{2D}}$, and is utilized to make predictions for atom and edge features by leveraging the underlying graph structure.

We enhance the MUformer's attention mechanism by incorporating pair-wise information in the invariant channel. 
First, we calculate intermediate representations for the edge and graph features
\begin{equation}
    \begin{aligned}
        &\mathbf{Z_{E_1}} = W_{\mathbf{E_1}}\mathbf{Z_E}, \ \mathbf{Z_{E_2}} = W_{\mathbf{E_2}}\mathbf{Z_E} \\
        &\mathbf{Z_{M_1}} = W_{\mathbf{M_1}}\mathbf{Z_M}, \ \mathbf{Z_{M_2}} = W_{\mathbf{M_2}}\mathbf{Z_M}, \ 
    \end{aligned}
\end{equation}
using weight matrices $W_{\mathbf{E_1}}, W_{\mathbf{E_2}}, W_{\mathbf{M_1}}, W_{\mathbf{M_2}}$. We then compute the attention weights by taking the dot product of the query and key, and modify them by multiplying and adding the intermediate representations,
\begin{equation}
    \begin{aligned}
       &\mathbf{A} = \frac{(W_Q\mathbf{Z_H})^T(W_K\mathbf{Z_H})}{\sqrt{F}} \in \mathbb{R}^{n\times n\times F} \\
        &\mathbf{A} = \mathbf{A} \times (\mathbf{Z_{E_1}+1)} + \mathbf{Z_{E_2}}.
    \end{aligned}
\end{equation}
And the predicted edge and graph representations, $\mathbf{\hat{Z}_E}, \mathbf{\hat{Z}_M}$, are computed from the attention weights as
\begin{equation}
    \begin{aligned}
       &\mathbf{\hat{Z}_E} = W_{\mathbf{E}_{\text{out}}}((\mathbf{A} \times (\mathbf{Z_{M_1}+1)} + \mathbf{Z_{M_2}}) \\
        &\mathbf{\hat{Z}_M} = W_{\mathbf{M}_{\text{out}}}(f_{\text{Node2Graph}}(\mathbf{Z_H})+ f_{\text{Edge2Graph}}(\mathbf{Z_E})+ \mathbf{{Z}_M}),
    \end{aligned}
\end{equation}
where $f_{\text{Node2Graph}}(\cdot)$ and $f_{\text{Edge2Graph}}(\cdot)$ are designated functions (see App~\ref{app:n2g.e2g}) that map node and edge features to graph features, respectively.  
Finally, the spatial relationships in 2D and 3D are added to the attention weights, the attention is passed through a softmax function and the predicted representation $ \mathbf{\hat{Z}_F}$ is obtained by the equation,
\begin{equation}
    \begin{aligned}
       & \mathbf{A} = \text{softmax}(\mathbf{A} + \Phi^{\text{2D}}_{\mathbf{E}} + \Phi^{\text{3D}}_{\mathbf{E}}) \in \mathbb{R}^{n\times n \times F}\\
        & \mathbf{\hat{Z}_H} = \mathbf{{Z}_H} + W_{\mathbf{H}_\text{out}}((W_V\mathbf{Z_H})\mathbf{A}).
    \end{aligned}
\end{equation}
The invariant channel can capture the inherent features of the input molecule graph, allowing for predictions of discrete 2D structures, as well as invariant atom and graph features.

\paragraph{Equivariant Channel}
The equivariant channel is an upgraded version of the one presented in \cite{tholke2022torchmd}. It is specifically engineered to extract the features of the input molecule graph $\mathbf{M}^{\text{3D}}$ that change under 3D rotations and translations. This channel is used to make predictions for atom features and coordinates by leveraging the 3D geometric structure of the molecule. It is activated when only the 3D geometric structure $\mathbf{M}^{\text{3D}}$ is provided. The velocity features ${\vv}\in \mathbb{R}^{n\times 3 \times F}$ are initialized to $0$.

First, we calculate the distance between each pair of atoms, $d_{ij}$, and project them into a multidimensional filter for keys. The attention weights are calculated by taking the dot product of the query, key, and filter, and are modified by incorporating the 3D spatial relationship between atoms. The attention weights are then passed through a softmax function, and the cosine cutoff is applied to the weights to ensure that atoms with a distance larger than $d_\text{cut}$ do not interact.
\begin{equation}
    \begin{aligned}
       & {d}_{ij} = \|{x}_i - {x}_j\|_2 \\
       & D_{K} = \text{SiLU}(W_{\text{dist}_K}(f_{\text{RBF}_3}(d)) + b_{\text{dist}_K}) \in \mathbb{R}^{n\times n \times F}\\
       & \mathbf{A} = \frac{(W_Q\mathbf{Z_H})^T(W_K\mathbf{Z_H})\odot D_{K}}{\sqrt{F}} \\ 
       & \mathbf{A} = \text{softmax}(\mathbf{A}+\Phi^{\text{3D}}_{\mathbf{E}}) \odot f_{\cos}(d).
    \end{aligned}
\end{equation}
Here, the final attention weights can also include 2D spatial relationship by adding 2D spatial relationship term $\Phi^{\text{2D}}_{\mathbf{E}}$ to the equation, and setting $f_{\cos}(d_{ij})=1$ if an edge exists between atoms $i$ and $j$ in the 2D molecule graph $\mathbf{M}^{\text{2D}}$.

In order to incorporate interatomic distances into the features directly, we also project the distance between atoms into a multidimensional filter for values. This approach, which has been used in \cite{schutt2017schnet, tholke2022torchmd}, enables the model to not only consider interatomic distances in the attention weights, but also to incorporate this information into the features themselves.
\begin{equation}
    \begin{aligned}
        & D_{V} = \text{SiLU}(W_{\text{dist}_V}(f_{\text{RBF}_3}(d)) + b_{\text{dist}_V}) \in \mathbb{R}^{n\times n\times 3F}\\
       & \mathbf{Z}_{V} = W_{V}\mathbf{Z_H} \in \mathbb{R}^{n \times 3F} \\
       & \mathbf{Z}_{V_1},\mathbf{Z}_{V_2},\mathbf{Z}_{V_3} = \text{split}(\mathbf{Z}_V \odot D_V) \in \mathbb{R}^{n\times n\times F} \\
        & \mathbf{Z}_{O} = W_{O}(\mathbf{Z}_{V_1}\mathbf{A}) \in \mathbb{R}^{n \times 3F} \\
       & \mathbf{Z}_{O_1},\mathbf{Z}_{O_2},\mathbf{Z}_{O_3} = \text{split}(\mathbf{Z}_O) \in \mathbb{R}^{n\times F},
    \end{aligned}
\end{equation}
where the function split$(\cdot)$ divides the input into three equal-sized parts.

Then, we use a weight matrix $W_{{\vv}}$ to project the velocity features ${\vv}$ into three separate vectors,
\begin{equation}
    \begin{aligned}
       & \mathbf{Z}_{{\vv}} = W_{{\vv}}{\vv} \in \mathbb{R}^{n\times 3 \times 3F} \\
       & \mathbf{Z}_{{{\vv}}_1},\mathbf{Z}_{{{\vv}}_2}, \mathbf{Z}_{{{\vv}}_3} = \text{split}(\mathbf{Z}_{{\vv}}) \in \mathbb{R}^{n\times 3\times F}.
    \end{aligned}
\end{equation}

Finally, new atom and velocity features, $\mathbf{\hat{Z}_H},\hat{{\vv}}$, are calculated following the steps in \cite{tholke2022torchmd}. The atom features are updated by adding the residual of the scaled features $\mathbf{Z}_{O_1}$ and the inner product between velocity projections $\langle \mathbf{Z}_{{{\vv}}_1}, \mathbf{Z}_{{{\vv}}_2} \rangle$. The velocity features are updated by incorporating equivariant features using the edge directional information $d_{ij}$ and scaled vector features.
\begin{equation}
    \begin{aligned}
       & \mathbf{\hat{Z}}_{\mathbf{H}} = \mathbf{Z_H} + (\mathbf{Z}_{O_1} + \mathbf{Z}_{O_2} \odot \langle  \mathbf{Z}_{{{\vv}}_1} , \mathbf{Z}_{{{\vv}}_2} \rangle) \\
       & {\vw}_i = \sum_{j \in N(i)} \mathbf{Z}_{V_{2, ij}} \odot {\vv}_i + \mathbf{Z}_{V_{3, ij}} \odot d_{ij} \\
       & \hat{{\vv}} =  {\vv} + ({\vw} + \mathbf{Z}_{O_3} \odot \mathbf{Z}_{{{\vv}}_3}).
    \end{aligned}
\end{equation}

\noindent \textbf{Interaction Embedding} \quad
We denote the predicted atom features of the invariant channel and the equivariant channel as $\mathbf{Z}_{\mathbf{H}}^{\text{inv}}$ and $\mathbf{Z}_{\mathbf{H}}^{\text{eqv}}$, respectively. We combine these predictions by multiplying them with a weight matrix ${W}_{\text{comb}_3}$, and adding a bias term ${b}_{\text{comb}_3}$,
\begin{equation}
    \mathbf{\hat{Z}_H} = {W}_{\text{comb}_3}[\mathbf{Z}_{\mathbf{H}}^{\text{inv}}, \mathbf{Z}_{\mathbf{H}}^{\text{eqv}}] + {b}_{\text{comb}_3}.
\end{equation}
By doing so, we obtain a mixed feature that includes rich invariant representations. This mixed atom feature $\mathbf{\hat{Z}_H}$ is then fed into the next layer of the transformer channels or used as input for the final predictions.

\noindent \textbf{Output Block} \quad
The output block generates the final output by utilizing the embeddings from invariant and equivariant channels. Specifically, it takes in the atom $\hat{\mathbf{Z}}_{\mathbf{H}}$ and edge embeddings $\hat{\mathbf{Z}}_{\mathbf{E}}$, along with the velocity embedding $\hat{{\vv}}$. Through feature extractions, the output block makes predictions for the atom features $\mathbf{H}_{\text{out}}$, edge features $\mathbf{E}_{\text{out}}$, and atom coordinates $\mathbf{X}_{\text{out}}$,
\begin{equation}
\begin{aligned}
    & \mathbf{H}_{\text{out}} = W_{{\mathbf{X}}_{\text{out}_2}}\left(\text{SiLU}(W_{{\mathbf{X}}_{\text{out}_1}}\hat{\mathbf{Z}}_{\mathbf{H}})\right) \in \mathbb{R}^{n\times d_{\text{out}}}\\
    & \mathbf{E}_{\text{out}} = W_{{\mathbf{E}}_{\text{out}_2}}\left(\text{SiLU}(W_{{\mathbf{E}}_{\text{out}_1}}\hat{\mathbf{Z}}_{\mathbf{E}})\right) \in \mathbb{R}^{n\times n \times b_{\text{out}}}\\
    & \mathbf{X}_{\text{out}} = \mathbf{X} + W_{{{\vv}}_{\text{out}_2}}\left(\text{SiLU}(W_{{{\vv}}_{\text{out}_1}}\hat{{\vv}})\right) \in \mathbb{R}^{n\times 3}.
\end{aligned} 
\end{equation}

\noindent \textbf{Analysis of Memory Complexity} \quad
We compare the memory complexity of our method, MUDiff, with two existing methods: EDM~\cite{hoogeboom2022equivariant} and DiGress~\cite{vignac2022digress}. Considering atom features of size $n \times d$, atom positions of size $n \times 3$, and edge features of size $n \times n \times b$, where $n$ is the number of atoms, $d$ is the dimension of atom features, and $b$ is the dimension of edge features, EDM's memory complexity is $O(nd + 3n)$, and DiGress's is $O(nd + n^2b)$. MUDiff has a higher memory complexity of $O(nd + 3n + n^2b)$, but offers a more comprehensive molecular representation by including both 2D and 3D information for topological and geometric structures. For more on scalability issues and potential solutions, see App~\ref{app:discussion}.

Unlike Transformer-M \cite{luo2022one}, which also employs a two-channel architecture to process 2D and 3D molecular data but is limited to predicting atom features only, our model can make predictions for atom features, molecular structures, and atom positions, while also being robust to geometric transformations. The two channels in Transformer-M are not explicitly designed as invariant and equivariant channels, which makes their model less robust to geometric transformations.

Our MUformer architecture can be used under different input conditions. When only 2D molecular information is available, only the invariant channel is activated, and the model makes predictions for atom and edge features only. When only 3D molecular information is available, only the equivariant channel is activated, and the model makes predictions for atom features and coordinates only. When both 2D and 3D molecular data are provided, the invariant and equivariant channels are activated, and the model can make predictions for the complete molecule, including atom features, molecular structure, and geometric structure. More details about our MUformer architecture, as well as a comparison with Transformer-M, can be found in App~\ref{app:transformer.arch}.

\section{Importance of 2D and 3D Attention Biases}
\label{app:discussion.attention.bias}
In Sec~\ref{sec:attention_biases}, we introduce two attention biases employed in MUformer to compute attentions, one for 2D molecular structure and another for 3D geometric structure. For the 2D spatial attention bias, we use the Shortest Path Distance (SPD) encoding to capture the distance between atoms $i$ and $j$ in the 2D molecular graph, providing vital structural relationship information. Additionally, we incorporate edge-type information along the shortest path, which reflects the connections between atoms $i$ and $j$, further enhancing the model's understanding of the 2D molecular graph. In the case of the 3D spatial attention bias, we calculate the bias using the Euclidean distance and an exponential radial basis function, which encodes the 3D spatial relationships between atom pairs in the 3D molecular structure. This approach helps the model account for the geometric arrangement of atoms in the molecule.
These biases are used to improve the model's representation of molecular structures by capturing essential structural features and relationships in both 2D and 3D spaces.

\paragraph{2D Attention Bias} The 2D spatial attention bias, which consists of the Shortest Path Distance (SPD) encoding and edge-type information, helps the model to capture the topological relationships between atoms in the 2D molecular graph. By incorporating this bias into the attention computation, the model can better understand the structural relationships and chemical properties of the molecule, leading to improved prediction and generation tasks.

\paragraph{3D Attention Bias} The 3D spatial attention bias encodes the 3D spatial relationships between atom pairs in the molecular structure using Euclidean distance and an exponential radial basis function. By including this information as a bias in the attention computation, the model can account for the geometric arrangement of atoms in the molecule. This allows the MUformer to recognize spatial patterns and interactions that are not apparent in the 2D graph representation alone.

 The attention biases introduced in Sec~\ref{sec:attention_biases} for both 2D and 3D molecular structures enhance the MUformer ability to capture essential structural features and relationships in both spaces. By incorporating these biases in the attention computation, the model can prioritize and focus on the most relevant connections between atoms, resulting in a more accurate and comprehensive molecular representation.

\section{Analysis of Interdependence of Generation of 2D and 3D Structures}
\label{app:discussion.2d3d.relation}

\paragraph{Motivation} Generating both 2D and 3D structures can provide a more complete representation of the molecular structure, as it captures both the planar arrangement of atoms in the molecule and their spatial arrangement in 3D space. By combining the generation of 2D and 3D structures, we can provide a more comprehensive understanding of the molecular structure, which could be useful for a variety of applications in drug discovery, materials science, and other fields. 

We discuss their effects from three perspectives, (1) conformational space, (2) stereochemistry, and (3) constraint.
\paragraph{Impact of 2D Generation on 3D Generation} 
(1) From the conformational perspective, the 2D structure can provide information about the planar arrangement of atoms in the molecule, which can be used to guide the generation of the 3D structure. By considering the 2D structure, the generation algorithm can explore different conformations and orientations of the molecule in a 3D space, which can help generate a more accurate 3D structure. 
(2) From the stereochemistry perspective, the 2D structure can provide information about the stereochemistry and chirality of the molecule, which can be used to guide the generation of the 3D structure. For example, if the 2D structure indicates that two atoms are connected by a double bond, the generation algorithm can infer the correct geometry for the double bond in the 3D space based on the stereochemistry of the molecule.
(3) From the constraints' perspective, the 2D structure can provide constraints on the geometry of the molecule, which can be used to guide the generation of the 3D structure. For example, if the 2D structure indicates that two atoms are connected by a ring, the generation algorithm can use this information to constrain the geometry of the ring in a 3D space.

\paragraph{Impact of 3D Generation on 2D Generation}
(1) From the conformational perspective, the generation of 3D structures can provide information about the conformational space of the molecule, which can be used to refine the 2D structure. For example, if the 3D structure indicates that two atoms are in close proximity, the generation algorithm can adjust the 2D structure to reflect this.
(2) From the stereochemistry perspective, the 3D structure can provide additional information about the stereochemistry and chirality of the molecule, which can be used to refine the 2D structure. For example, if the 3D structure indicates that two atoms have a specific orientation in 3D space, the generation algorithm can adjust the 2D structure to reflect this.
(3) From the constraint perspective, the 3D structure can provide additional constraints on the geometry of the molecule, which can be used to refine the 2D structure. For example, if the 3D structure indicates that two atoms are connected by a ring, the generation algorithm can adjust the 2D structure to ensure that the ring is planar.

\section{Additional Empirical Results}
\label{app:additional.experiment}

\subsection{Molecule Generation on DRUG}
\label{app:molecule.generation.drug}
We compare the performance of our MUDiff 
model with popular generative models, including  GDM \cite{hoogeboom2022equivariant}, EDM \cite{hoogeboom2022equivariant}, MDM \cite{huang2023mdm}, and GeoLDM \cite{xu2023geometric}. The results are reported in Table~\ref{tab:mol.generation.drug} 
As outlined in \cite{huang2023mdm, xu2023geometric}, we evaluate the atom and molecule stability of the generated compounds.
\begin{table*}[htbp!]
  \centering
  \caption{Atom stability, molecule stability, and validity are evaluated across 3 runs on DRUG.}
  \resizebox{9.0cm}{!}{
  \footnotesize
    \begin{tabular}{lccc}
    \hline
    \hline
    \textbf{Method} & Validity & Atom Stable(\%) & Mol Stable(\%)\\
    \hline
    \textbf{GDM} & $90.8$ & $75.0 $ & $- $\\
    \textbf{EDM} & $92.6$ & $ 81.3 $ & $13.0 $\\
    \textbf{MDM} & 99.8 & $ - $  & $ 62.2 $\\
    \textbf{GeoLDM} & 99.3 & $84.4$  & $- $\\
    \textbf{MUDiff} (ours) & $ 98.9$  &   $84.0$    &  $60.9$\\
    \hline
    \hline
    \end{tabular}%
    }
  \label{tab:mol.generation.drug}%
\end{table*}

\subsection{Conditional Generation}
\label{app:conditional.generation.results}
Our method can be extended for conditional molecule generation, as detailed in App~\ref{app:uncond.generation}.
We follow the experimental setting in \cite{hoogeboom2022equivariant} to train the conditional MUDiff model on the QM9 dataset, conditioning the generation on properties $\alpha$, $\epsilon_{\text{homo}}$, $\epsilon_{\text{lumo}}$, $\Delta\epsilon$, $\mu$, and $C_v$, respectively. 

Additionally, we follow \cite{hoogeboom2022equivariant} to use a property classifier $\psi_c$ proposed in \cite{satorras2021n2}. We split QM9 training data into two halves, \textit{A} and \textit{B}, each containing 50K samples, and use \textit{A} subset to train $\psi_c$ and \textit{B} subset for training the conditional MUDiff. Then, $\psi_c$ is used to evaluate the generated samples of conditional MUDiff. Also, we follow \cite{hoogeboom2022equivariant} to report the loss of $\psi_c$ on \textit{B} as a lower bound (L-bound). 
The smaller the gap between MUDiff and L-bound, the more similar MUDiff generated samples to \textit{B}.

\begin{table*}[htbp!]
  \centering
  \caption{Mean Absolute Error for the prediction of molecular properties by the property classifier $\psi_c$ on a QM9 subset (L-bound), MUDiff samples and three baselines.}
  \resizebox{9.0cm}{!}{
  \footnotesize
    \begin{tabular}{lcccccc}
    \hline
    \hline
    \textbf{Property} & $\alpha$ & $\epsilon_{\text{homo}}$ & $\epsilon_{\text{lumo}}$ & $\Delta\epsilon$ & $\mu$ & $C_v$ \\
    \textbf{Units} & $a^3$ & $meV$ & $meV$ & $meV$ & $D$   & $meV$ \\
    \hline
    \textbf{U-bound} & 9.01  & 645   & 1457  & 1470  & 1.616 & 6.857 \\
    \textbf{\#Atoms} & 3.86  & 426   & 813   & 866   & 1.053 & 1.971 \\
    \textbf{EDM} & 2.76  & 356   & \textbf{584}   & 655   & 1.111 & 1.101 \\
    \textbf{MUDiff} (ours) &   \textbf{2.15}    &  \textbf{315}     &  597     &  \textbf{604}      &  \textbf{1.033}     &  \textbf{0.978} \\
    \textbf{L-bound} & 0.10   & 39    & 36    & 64    & 0.043 & 0.040 \\
    \hline
    \hline
    \end{tabular}%
    }
  \label{tab:cond.generation.MUDiff}%
\end{table*}

We evaluate the performance against the baselines used in EDM \cite{hoogeboom2022equivariant}. In addition to L-bound, they also use two other baselines: U-bound and \#Atoms. The U-bound is obtained by shuffling the properties of molecules in the \textit{B} subset and evaluating $\psi_c$ on it. The \#Atoms baseline predicts the molecular properties in the \textit{B} subset by only using the number of atoms in the molecule.

\paragraph{Results}
Table~\ref{tab:cond.generation.MUDiff} showcases the results of conditional generation on the QM9 dataset. Evidently, conditional MUDiff generates samples that more closely resemble the molecules in subset B compared to the baselines, indicating that MUDiff outperforms baselines in generating molecules with desired properties and capturing structural similarities.

\subsubsection{Conditional Generation}
\label{app:uncond.generation}
We follow the conditional molecule generation procedure in \cite{hoogeboom2022equivariant}. We train the conditional MUDiff by concatenating the conditions $c$ with atom features, the Eq~\ref{eq:uncond} is rewritten as
\begin{equation}
\begin{aligned}
\mathbf{\hat{{\bm{\epsilon}}}}^t_{\mathbf{H}}, p(\mathbf{\hat{E}}), \mathbf{\hat{{\bm{\epsilon}}}}^t_{\mathbf{X}}=
\psi_{\theta}([\mathbf{\tilde{H}}_t, \frac{t}{T}, c], \mathbf{\tilde{E}}_t, \mathbf{\tilde{X}}_t) - (\mathbf{0}, \mathbf{0}, \mathbf{\tilde{X}}_t),
\end{aligned}
\end{equation}
and the loglikelihood (in Eq~\ref{eq:likelihood}) is modified to
\begin{equation}
\begin{aligned}
\log p(\mathbf{M} | c) \geq 	\underbrace{\KL[q(\mathbf{M}_T | \mathbf{M}, c)\| p(\mathbf{M}_T | c) ]}_{\text{Prior loss}} + \underbrace{\mathbb{E}_{q(\mathbf{M}_0|\mathbf{M}, c)}[\log p(\mathbf{M}|\mathbf{M}_0, c)]}_{\text{Reconstruction loss}} + \underbrace{\sum_{t=1}^T\mathcal{L}_{t,c}}_{\text{Diffusion loss}},
\end{aligned}
\end{equation}
where
\begin{equation}
\begin{aligned}
\mathcal{L}_{t,c} = \mathbb{E}_{q(\mathbf{M}_t|\mathbf{M}, c)}[\KL [q(\mathbf{M}_{t-1}|\mathbf{M}_t, \mathbf{M}, c)\|p(\mathbf{M}_{t-1}|\mathbf{M}_t, c)]].
\end{aligned}
\end{equation}

In order to perform conditional sampling, we adopt the method outlined in \cite{hoogeboom2022equivariant}. Specifically, we first sample a condition $c$ from the condition distribution, and then use this condition to generate molecules via the conditional distribution $p(\mathbf{\hat{M}} | c)$. The details of the model training procedure and architecture can be found in App~\ref{app:molecule.generation}. This approach allows us to generate molecules that are conditioned on specific properties, such as electronic energy levels or heat capacity.

\paragraph{Properties} Following the methodology used in previous works such as \cite{hoogeboom2022equivariant}, we condition the generation of molecules on several properties. Specifically, we condition the generation on the polarizability ($\alpha$), the highest occupied molecular orbital energy ($\epsilon_{\text{homo}}$), the lowest unoccupied molecular orbital energy ($\epsilon_{\text{lumo}}$), the energy difference between the highest occupied and lowest unoccupied molecular orbital energy ($\Delta\epsilon$), the dipole moment ($\mu$) and the heat capacity at 298.15K ($C_v$). These properties provide specific information about the physical and chemical properties of a molecule, allowing us to generate molecules with specific desired characteristics.

\subsection{Property Prediction on QM9}
\label{app:qm9.property}
We conduct a thorough comparison of our MUformer model against several state-of-the-art (SOTA) models for property prediction on the QM9 dataset, including SchNet \cite{schutt2017schnet}, EGNN \cite{satorras2021n}, PhysNet \cite{unke2019physnet}, DimeNet \cite{beani2021directional}, Cormorant \cite{anderson2019cormorant}, PaiNN \cite{schutt2021equivariant}, and ET \cite{tholke2022torchmd}. The results, which can be found in Table~\ref{tab:qm9.property}, are obtained by averaging over three runs. We use a learning rate of 1e-3, 1e-4, 5e-4 and 1e-5, weight decay of 5e-5, 128 hidden dimensions, and 6 layers for our MUformer model. The results of the baseline models are taken from \cite{tholke2022torchmd}.

\begin{table*}[htbp!]
  \centering
  \caption{Results on all QM9 targets and comparison to previous literature. Scores are reported as mean absolute errors (MAE) with standard deviation. Results of different models are averaged over three runs.}
  \resizebox{\textwidth}{!}{
  \footnotesize
    \begin{tabular}{llcccccccc}
    \hline
    \hline
    \multicolumn{1}{c}{Target} & \multicolumn{1}{c}{Unit} & SchNet & EGNN  & PhysNet & DimeNet++ & Cormorant & PaiNN & ET    & MUformer \\
    \hline
    $\mu$ & $D$   & 0.033 & 0.029 & 0.0529 & 0.041 & 0.0297 & 0.012 & {0.011} & 0.013 $\pm$ 0.003\\
    $\alpha$ & $a^3_0$ & 0.235 & 0.071 & 0.0615 & {0.0435} & 0.085 & 0.045 & 0.059 & 0.041 $\pm$ 0.008 \\
    $\epsilon_{\text{HOMO}}$ & $meV$   & 41    & 29    & 32.9  & 24.6  & 34    & 27.6  & 20.3  & 24.7 $\pm$ 1.2 \\
    $\epsilon_{\text{LUMO}}$ & $meV$   & 34    & 25    & 24.7  & 19.5  & 38    & 20.4  & 17.5  & 	20.2 $\pm$ 0.8 \\
    $\Delta \epsilon$ & $meV$   & 63    & 48    & 42.5  & 32.6  & 61    & 45.7  & 36.1  & 30.3 $\pm$ 1.7 \\
    \textlangle$R^2$\textrangle     & $a^2_0$ & 0.073 & 0.106 & 0.765 & 0.331 & 0.961 & 0.066 & {0.033} & 0.117 $\pm$ 0.012 \\
    $ZPVE$  & $meV$   & 1.7   & 1.55  & 1.39  & {1.21}  & 2.027 & 1.28  & 1.84  & 1.76  $\pm$ 0.08 \\
    $\mathbf{U}_0$   & $meV$   & 14    & 11    & 8.15  & 6.32  & 22    & 5.85  & 6.15  & 6.11 $\pm$ 0.12 \\
    $\mathbf{U}$      & $meV$   & 19    & 12    & 8.34  & 6.28  & 21    & 5.83  & 6.38  & 6.04 $\pm$ 0.19 \\
    $\mathbf{H}$     & $meV$   & 14    & 12    & 8.42  & 6.53  & 21    & {5.98}  & 6.16  &  6.77 $\pm$ 0.07 \\
    $\mathbf{G}$      & $meV$   & 14    & 12    & 9.4   & 7.56  & 20    & 7.35  & 7.62  &  7.24 $\pm$ 0.08 \\
    $\mathbf{C}_v$   & $\frac{cal}{mol\ K}$ & 0.033 & 0.031 & 0.028 & {0.023} & 0.026 & 0.024 & 0.026 & 0.023 $\pm$ 0.002 \\
    \hline
    \hline
    \end{tabular}%
    }
  \label{tab:qm9.property}%
\end{table*}

\subsection{Ablation Study}
\label{app:ablation.study}
To investigate how different components will impact the performance of MUDiff, we conduct ablation study in this subsection . The details of the experimental settings can be found in App~\ref{app:ablation}.

\begin{table*}[htbp!]
  \centering
  \caption{The performance of molecule stability and validity across different ablation models, as shown by the average value with standard deviation of 1K generated molecules (with hydrogen) from each model. Model variations include using $(\romannumeral 1)$ 2D structure encoding, $(\romannumeral 2)$ graph encoding, $(\romannumeral 3)$ 2D neighborhood encoding, $(\romannumeral 4)$ 3D neighborhood encoding, $(\romannumeral 5)$ 2D spatial attention bias, $(\romannumeral 6)$ 3D spatial attention bias, $(\romannumeral 7)$ edge features as attention bias, $(\romannumeral 8)$ graph features as attention bias, and $(\romannumeral 9)$ 2D discrete graph structures into 3D geometric structures.}
   \resizebox{10.0cm}{!}{
  \footnotesize
    \begin{tabular}{llll|llll|lcc}
    \hline
    \hline
    \multicolumn{4}{c|}{\textbf{Encoding}} & \multicolumn{4}{c|}{\textbf{Bias}} &       &       &  \\
    \hline
    $\romannumeral 1$     &  $\romannumeral 2$    &  $\romannumeral 3$    &  $\romannumeral 4$   &  $\romannumeral 5$   &  $\romannumeral 6$    &  $\romannumeral 7$    &  $\romannumeral 8$   &  $\romannumeral 9$   & Mol Stable (\%) & Valid (\%) \\
    \hline
          &       &       &       &       &       &       &       &       & $82.5 \pm 6.3$     & $90.7 \pm 2.2$ \\
    \checkmark &       &       &       &       &       &       &       &       & $82.7 \pm 1.9$     & $91.3 \pm 1.7$ \\
    \checkmark &       & \checkmark &       &       &       &       &       &       & $82.9 \pm 1.5$     & $91.6 \pm 2.1$ \\
    \checkmark &       & \checkmark &       & \checkmark &       & \checkmark &       &       & $84.7 \pm 1.2$     & $92.9 \pm 1.3$ \\
    \checkmark &       &       & \checkmark &       &       &       &       &       & $85.3 \pm 1.0$     & $93.6 \pm 1.9$ \\
    \checkmark &       &       & \checkmark &       & \checkmark &       &       &       & $86.1 \pm 1.7$     & $93.5 \pm 1.3$ \\
    \checkmark & \checkmark & \checkmark & \checkmark &  &       &  &       &       &  $87.2 \pm 1.6$     & $93.4 \pm 1.0$ \\
    \checkmark & \checkmark & \checkmark & \checkmark &  &  \checkmark     &  &       &       &  $87.1 \pm 1.5$     & $94.8 \pm 1.4$ \\
    \checkmark & \checkmark & \checkmark & \checkmark &  \checkmark   & \checkmark &       &       &       & $88.3 \pm 1.5$     & $94.5 \pm 1.3$ \\
    \checkmark & \checkmark & \checkmark & \checkmark & \checkmark & \checkmark & \checkmark & \checkmark &       & $88.4 \pm 1.6$     & $95.1 \pm 1.3$ \\
    \checkmark & \checkmark & \checkmark & \checkmark & \checkmark & \checkmark & \checkmark & \checkmark & \checkmark & $89.3 \pm 1.3$     & $95.3 \pm 1.2$ \\
    \hline
    \hline
    \end{tabular}%
    }
  \label{tab:ablation}%
\end{table*}

\paragraph{Results}
The results in Table~\ref{tab:ablation} demonstrate the effectiveness of each component in MUformer. We can see that each component proposed in Sec~\ref{sec:graph.transformer} plays an indispensable role in learning all aspects of the molecule in a unified manner and the diffusion model with all components combined generates the most stable and valid molecules.

\subsubsection{Ablation Study}
\label{app:ablation}
To improve efficiency, we conduct ablation studies using smaller models than those described in App~\ref{app:molecule.generation}. Specifically, these models consist of 4 layers, 64 embedding dimensions for atom- and edge-level features, 32 embedding dimensions for graph-level features, 8 attention heads, 100 feedforward dimensions for atom- and edge-level features, 50 feedforward dimensions for graph-level features, 0.3 dropout rate for all latent embeddings and attention values, SiLU activation function, 1e-4 learning rate, and Adam optimizer with 5e-5 weight decay. Additionally, we use a diffusion process with 500 time steps over the course of 3000 training epochs.

Additionally, if 2D structure encoding is not used, the edge type $\mathbf{E}$ is simply embedded by a weight matrix $W$ as
\begin{equation}
\begin{aligned}
\mathbf{Z_E} = W\mathbf{E} \in \mathbb{R}^{n\times n \times fe_{\text{in}}}.
\end{aligned}
\end{equation}

To reduce computation costs, we sample 1,000 molecules from each MUDiff ablation model instead of the 10,000 samples typically generated by the diffusion model during sampling. This allows us to evaluate the performance of the various MUDiff models while minimizing the computational resources required. 

Model variations for ablation study include: (1) no extra technique, (2) 2D structure encoding, (3) 2D structure encoding + 2D neighborhood encoding, (4) 2D structure encoding + 2D neighborhood encoding + 2D spatial attention bias + edge features as attention bias, (5) 2D structure encoding + 3D neighborhood encoding, (6) 2D structure encoding + 3D neighborhood encoding + 3D spatial attention bias, (7) 2D structure encoding + graph encoding + 2D neighborhood encoding + 3D neighborhood encoding, (8) 2D structure encoding + graph encoding + 2D neighborhood encoding + 3D neighborhood encoding + 3D spatial attention bias, (9) 2D structure encoding + graph encoding + 2D neighborhood encoding + 3D neighborhood encoding + 2D spatial attention bias + 3D spatial attention bias, (10) 2D structure encoding + graph encoding + 2D neighborhood encoding + 3D neighborhood encoding + 2D spatial attention bias + 3D spatial attention bias + edge features as attention bias + graph features as attention bias, (11) 2D structure encoding + graph encoding + 2D neighborhood encoding + 3D neighborhood encoding + 2D spatial attention bias + 3D spatial attention bias + edge features as attention bias + graph features as attention bias + 2D discrete graph structures into 3D geometric structures.

\section{Limit Distribution for Edges}
\label{app:limit.edge}
\cite{vignac2022digress} suggests that the limit distribution $q_{\infty}=\lim_{T\to \infty}q(\mathbf{\Tilde{E}}_T|\mathbf{E})$
should be independent of clean data $\mathbf{E}$ for efficient diffusion models. In our diffusion model, the discrete process for noising/denoising discrete graph structures $\mathbf{E}\in \mathbb{R}^{n\times n\times b}$, we use a sequence of transition matrices $\{\bar{Q}_t\}_{t=0}^T$ to add noise to $\mathbf{E}$. In our choice, we follow \cite{austin2021structured, vignac2022digress} to use the simplest uniform transition parameterized by
\begin{equation}
\begin{aligned}
&\bar{Q}_t = \alpha_t \mathbf{I} + (1-\alpha_t)\frac{\mathbbm{1}_b\mathbbm{1}^T_b}{b} \in \mathbb{R}^{b\times b}\\ 
&\mathbf{\Tilde{E}}_{t} = \mathbf{E}\bar{Q}_t \in \mathbb{R}^{n\times n\times b}
\end{aligned}
\end{equation}
with $\alpha_t$ smoothly transition from $1\to 0$ as $t$ goes from $0\to T$. When $t$ gradually goes to $\infty$,
\begin{equation}
\begin{aligned}
\lim_{t\to \infty} \Bar{Q}_t &= \lim_{t\to \infty} \alpha_t \mathbf{I} + (1-\alpha_t)\frac{\mathbbm{1}_b\mathbbm{1}^T_b}{b} \\
&= \lim_{t\to \infty} \alpha_t \mathbf{I} + (1- \lim_{t\to \infty} \alpha_t)\frac{\mathbbm{1}_b\mathbbm{1}^T_b}{b} \\
&= \frac{\mathbbm{1}_b\mathbbm{1}^T_b}{b}.
\end{aligned}
\end{equation}
It suggests that $q(\mathbf{\Tilde{E}}_t|\mathbf{E})$ converges to a uniform distribution as $t\to T$, and the limit distribution $q_{\infty}$ is just a uniform distribution over the edge categories independently of $\mathbf{E}$.

\section{Likelihood and Lower Bound}
\label{app:lower.bound}
For simplicity, we use $\mathbf{M}_t=(\mathbf{H_t,E_t,X_t})$ to denote the noisy molecule $\mathbf{\Tilde{M}}_t=(\mathbf{\Tilde{H}}_t,\mathbf{\Tilde{E}}_t,\mathbf{\Tilde{X}}_t)$ at time $t$.
Following \cite{kingma2021variational}, the variational lower bound on the log-likelihood of a molecule $\mathbf{M}$ is derived as 
\begin{equation}
\label{eq:likelihood}
\log p(\mathbf{M}) \geq 	\underbrace{\KL[q(\mathbf{M}_T | \mathbf{M})\| p(\mathbf{M}_T) ]}_{\text{Prior loss}} + \underbrace{\mathbb{E}_{q(\mathbf{M}_0|\mathbf{M})}[\log p(\mathbf{M}|\mathbf{M}_0)]}_{\text{Reconstruction loss}} + \underbrace{\sum_{t=1}^T\mathcal{L}_t}_{\text{Diffusion loss}},
\end{equation}
where
\begin{equation}
\mathcal{L}_t = \mathbb{E}_{q(\mathbf{M}_t|\mathbf{M})}[\KL [q(\mathbf{M}_{t-1}|\mathbf{M}_t, \mathbf{M})\|p(\mathbf{M}_{t-1}|\mathbf{M}_t)]].
\end{equation}
The prior loss is fairly simple, 
\begin{equation}
\begin{aligned}
    \KL[q(\mathbf{M}_T | \mathbf{M})\| p(\mathbf{M}_T) ] & = \KL[q(\mathbf{H}_T, \mathbf{E}_T, \mathbf{X}_T | \mathbf{H}, \mathbf{E}, \mathbf{X})\| p(\mathbf{H}_T, \mathbf{E}_T, \mathbf{X}_T)  ] \\
    &\textit{by chain rule} \\
    & = \KL [q(\mathbf{H}_T|\mathbf{H,E,X}) \| p(\mathbf{H}_T)] + \KL [q(\mathbf{E}_T, \mathbf{X}_T| \mathbf{H,E,X}, \mathbf{H}_T) \| p(\mathbf{E}_T, \mathbf{X}_T | \mathbf{H}_T)]\\
    & = \KL [q(\mathbf{H}_T|\mathbf{H,E,X}) \| p(\mathbf{H}_T)] + \KL [q(\mathbf{E}_T| \mathbf{H,E,X}, \mathbf{H}_T) \| p(\mathbf{E}_T| \mathbf{H}_T)] \\
    & + \KL [q(\mathbf{X}_T| \mathbf{H,E,X}, \mathbf{H}_T, \mathbf{E}_T) \| p(\mathbf{X}_T | \mathbf{H}_T, \mathbf{E}_T))] \\
    &\textit{due to indenpendence} \\
    & = \KL[q(\mathbf{H}_T | \mathbf{H}) \| p(\mathbf{H}_T)] + \KL[q(\mathbf{E}_T | \mathbf{E}) \| p(\mathbf{E}_T)] + \KL[q(\mathbf{X}_T | \mathbf{X}) \| p(\mathbf{X}_T)],
\end{aligned}
\end{equation}
where $\KL[q(\mathbf{H}_T | \mathbf{H}) \| p(\mathbf{H}_T)]$ models the distance between a standard normal distribution $\mathcal{N}_{\mathbf{H}}(\mathbf{0, I})$ and the final noisy variable $q(\mathbf{H}_T|\mathbf{H})$, $\KL[q(\mathbf{X}_T | \mathbf{X}) \| p(\mathbf{X}_T)]$ models the distance between another standard normal distribution $\mathcal{N}_{\mathbf{X}}(\mathbf{0, I})$ and the final noisy variable $q(\mathbf{X}_T|\mathbf{X})$, and $\KL[q(\mathbf{E}_T | \mathbf{E}) \| p(\mathbf{E}_T)]$ measures the KL divergence between the uniform distribution over edge categories and the final noisy variable $q(\mathbf{E}_T | \mathbf{E})$. The prior loss $\KL[q(\mathbf{M}_T | \mathbf{M})\| p(\mathbf{M}_T) ]$ is always close to zero.

The reconstruction loss,
\begin{equation}
\begin{aligned}
        \mathbb{E}_{q(\mathbf{M}_0|\mathbf{M})}[\log p(\mathbf{M}|\mathbf{M}_0)] & = \mathbb{E}_{q(\mathbf{H}_0, \mathbf{E}_0, \mathbf{X}_0|\mathbf{H, E, X})}[\log p(\mathbf{H, X, E}|\mathbf{H}_0, \mathbf{E}_0, \mathbf{X}_0)] \\
        & =  \mathbb{E}_{q(\mathbf{H}_0, \mathbf{E}_0, \mathbf{X}_0|\mathbf{H, E, X})}[\log p(\mathbf{H}|\mathbf{H}_0) p(\mathbf{E}|\mathbf{E}_0) p(\mathbf{X}|\mathbf{X}_0)] \\
        & =  \mathbb{E}_{q(\mathbf{H}_0, \mathbf{E}_0, \mathbf{X}_0|\mathbf{H, E, X})}[\log p(\mathbf{H}|\mathbf{H}_0) + \log p(\mathbf{E}|\mathbf{E}_0) + \log p(\mathbf{X}|\mathbf{X}_0)] \\
        & =  \mathbb{E}_{q(\mathbf{H}_0|\mathbf{H,E,X})}[\log p(\mathbf{H}|\mathbf{H}_0)]  \mathbb{E}_{q(\mathbf{E}_0, |\mathbf{H,E,X})}[\log p(\mathbf{E}|\mathbf{E}_0)]  \mathbb{E}_{q(\mathbf{X}_0|\mathbf{H,E,X})}[\log p(\mathbf{X}|\mathbf{X}_0)] \\
        & =  \mathbb{E}_{q(\mathbf{H}_0|\mathbf{H})}[\log p(\mathbf{H}|\mathbf{H}_0)]  \mathbb{E}_{q(\mathbf{E}_0, |\mathbf{E})}[\log p(\mathbf{E}|\mathbf{E}_0)]  \mathbb{E}_{q(\mathbf{X}_0|\mathbf{X})}[\log p(\mathbf{X}|\mathbf{X}_0)].
\end{aligned}
\end{equation}
For discrete data $\mathbf{E}$,  $\log p(\mathbf{E}|\mathbf{E}_0)$ is computed from the probability of clean edge features given noisy edge features $p(\mathbf{E}|\mathbf{E}_0)$. For continuous data $\mathbf{H, X}$,  $\log p(\mathbf{H}|\mathbf{H}_0)$ models the likelihood of $\mathbf{H}$ given $\mathbf{H}_0\sim q(\mathbf{H}_0|\mathbf{H})$, $\log p(\mathbf{X}|\mathbf{X}_0)$ models the likelihood of $\mathbf{X}$ given $\mathbf{X}_0\sim q(\mathbf{X}_0|\mathbf{X})$, and the details of zeroth likelihood is discussed in App~\ref{app:zeroth.likelihood}.

The diffusion loss is different than the training loss in Eq~\ref{eq:training.atom.loss},\ref{eq:training.edge.loss}, but still at each time step $t$, the diffusion loss is composed of atom feature loss, edge feature loss, and atom coordinate loss,
\begin{equation}
\begin{aligned}
    \mathcal{L}_t & = \mathbb{E}_{q(\mathbf{M}_t|\mathbf{M})}[\KL [q(\mathbf{M}_{t-1}|\mathbf{M}_t, \mathbf{M})\|p(\mathbf{M}_{t-1}|\mathbf{M}_t)]] \\
    & = \mathbb{E}_{q(\mathbf{M}_t|\mathbf{M})}[\KL [q(\mathbf{H}_{t-1}, \mathbf{E}_{t-1}, \mathbf{X}_{t-1}|\mathbf{M}_t, \mathbf{M})\|p(\mathbf{H}_{t-1}, \mathbf{E}_{t-1}, \mathbf{X}_{t-1}|\mathbf{M}_t)]] \\
    & \textit{by chain rule} \\
    & = \mathbb{E}_{q(\mathbf{M}_t|\mathbf{M})} [\KL[q(\mathbf{H}_{t-1}|\mathbf{M}_t, \mathbf{M}) \| p(\mathbf{H}_{t-1}|\mathbf{M}_t)] +  \KL[q(\mathbf{E}_{t-1}, \mathbf{X}_{t-1}|\mathbf{M}_t, \mathbf{M}, \mathbf{H}_{t-1}) \| p(\mathbf{E}_{t-1}, \mathbf{X}_{t-1}|\mathbf{M}_t, \mathbf{H}_{t-1})]] \\
    & = \mathbb{E}_{q(\mathbf{M}_t|\mathbf{M})} [\KL[q(\mathbf{H}_{t-1}|\mathbf{M}_t, \mathbf{M}) \| p(\mathbf{H}_{t-1}|\mathbf{M}_t)] +  \KL[q(\mathbf{E}_{t-1}|\mathbf{M}_t, \mathbf{M}, \mathbf{H}_{t-1}) \| p(\mathbf{E}_{t-1}|\mathbf{M}_t, \mathbf{H}_{t-1})] \\
    & + \KL[q(\mathbf{X}_{t-1}|\mathbf{M}_t, \mathbf{M}, \mathbf{H}_{t-1}, \mathbf{E}_{t-1}) \| p(\mathbf{X}_{t-1}|\mathbf{M}_t, \mathbf{H}_{t-1}, \mathbf{E}_{t-1})]] \\
    &\textit{due to independence} \\
    & = \mathbb{E}_{q(\mathbf{M}_t|\mathbf{M})} [\KL[q(\mathbf{H}_{t-1}|\mathbf{M}_t, \mathbf{M}) \| p(\mathbf{H}_{t-1}|\mathbf{M}_t)] +  \KL[q(\mathbf{E}_{t-1}|\mathbf{M}_t, \mathbf{M}) \| p(\mathbf{E}_{t-1}|\mathbf{M}_t)] + \KL[q(\mathbf{X}_{t-1}|\mathbf{M}_t, \mathbf{M}) \| p(\mathbf{X}_{t-1}|\mathbf{M}_t)]] \\
    & = \mathbb{E}_{q(\mathbf{M}_t|\mathbf{M})} [\KL[q(\mathbf{H}_{t-1}|\mathbf{M}_t, \mathbf{M}) \| p(\mathbf{H}_{t-1}|\mathbf{M}_t)]] + \mathbb{E}_{q(\mathbf{M}_t|\mathbf{M})} [\KL[q(\mathbf{E}_{t-1}|\mathbf{M}_t, \mathbf{M}) \| p(\mathbf{E}_{t-1}|\mathbf{M}_t)]] \\
    & + \mathbb{E}_{q(\mathbf{M}_t|\mathbf{M})} [\KL[q(\mathbf{X}_{t-1}|\mathbf{M}_t, \mathbf{M}) \| p(\mathbf{X}_{t-1}|\mathbf{M}_t)]] \\
    & = \mathbb{E}_{q(\mathbf{H}_t|\mathbf{H})} [\KL[q(\mathbf{H}_{t-1}|\mathbf{H}_t, \mathbf{H}) \| p(\mathbf{H}_{t-1}|\mathbf{H}_t)]] + \mathbb{E}_{q(\mathbf{E}_t|\mathbf{E})} [\KL[q(\mathbf{E}_{t-1}|\mathbf{E}_t, \mathbf{E}) \| p(\mathbf{E}_{t-1}|\mathbf{E}_t)]] \\
    & + \mathbb{E}_{q(\mathbf{X}_t|\mathbf{X})} [\KL[q(\mathbf{X}_{t-1}|\mathbf{X}_t, \mathbf{X}) \| p(\mathbf{X}_{t-1}|\mathbf{X}_t)]].
\end{aligned}
\end{equation}
For discrete data $\mathbf{E}$, $\KL[q(\mathbf{E}_{t-1}|\mathbf{E}_t, \mathbf{E}) \| p(\mathbf{E}_{t-1}|\mathbf{E}_t)]$ measures the KL divergence between the true categorical distribution $q(\mathbf{E}_{t-1}|\mathbf{E}_t, \mathbf{E})$ and the predicted categorical distribution $p(\mathbf{E}_{t-1}|\mathbf{E}_t)$. For continuous data $\mathbf{H,X}$ with Gaussian noise being added, \cite{kingma2021variational, hoogeboom2022equivariant} show that $\KL[q(\mathbf{H}_{t-1}|\mathbf{H}_t, \mathbf{H}) \| p(\mathbf{H}_{t-1}|\mathbf{H}_t)], \KL[q(\mathbf{X}_{t-1}|\mathbf{X}_t, \mathbf{X}) \| p(\mathbf{X}_{t-1}|\mathbf{X}_t)]$ can be expressed as,
\begin{equation}
\begin{aligned}
    \KL[q(\mathbf{H}_{t-1}|\mathbf{H}_t, \mathbf{H}) \| p(\mathbf{H}_{t-1}|\mathbf{H}_t)] &= \frac{1}{2} \mathbb{E}_{{\bm{\epsilon}}_{\mathbf{H}}^t \sim N_{\mathbf{H}}(\mathbf{0,I})}\left[ \omega(t) \| {\bm{\epsilon}}_{\mathbf{H}}^t-\mathbf{\hat{{\bm{\epsilon}}}}^t_{\mathbf{H}} \|^2\right] \\
    \KL[q(\mathbf{X}_{t-1}|\mathbf{X}_t, \mathbf{X}) \| p(\mathbf{X}_{t-1}|\mathbf{X}_t)] &= \frac{1}{2} \mathbb{E}_{{\bm{\epsilon}}_{\mathbf{X}}^t \sim N_{\mathbf{X}}(\mathbf{0,I})}\left[ \omega(t) \| {\bm{\epsilon}}_{\mathbf{X}}^t-\mathbf{\hat{{\bm{\epsilon}}}}^t_{\mathbf{X}} \|^2\right],
\end{aligned}
\end{equation}
where $\omega(t) = (1-\text{SNR}(t-1)/\text{SNR}(t))$.

\section{Posterior Distribution of Edge Features $p(\mathbf{\Tilde{E}}_{t-1}|\mathbf{\Tilde{E}}_t)$}
\label{app:posterior.edge}
For simplicity, we use $\mathbf{M}_t=(\mathbf{H_t,E_t,X_t})$ to denote the noisy molecule $\mathbf{\Tilde{M}}_t=(\mathbf{\Tilde{H}}_t,\mathbf{\Tilde{E}}_t,\mathbf{\Tilde{X}}_t)$ at time $t$.
The posterior distribution of a molecule is calculated by,
\begin{equation}
\begin{aligned}
    p(\mathbf{{M}}_{t-1}|\mathbf{{M}}_{t}) & = p(\mathbf{{H}}_{t-1}, \mathbf{{E}}_{t-1}, \mathbf{{X}}_{t-1} | \mathbf{{H}}_{t}, \mathbf{{E}}_{t}, \mathbf{{X}}_{t}) \quad\quad \textit{$\mathbf{H, E, X}$ are independent} \\ 
    & = p(\mathbf{{H}}_{t-1} | \mathbf{{H}}_{t})
    p(\mathbf{{E}}_{t-1} | \mathbf{{E}}_{t})
    p(\mathbf{{X}}_{t-1} | \mathbf{{X}}_{t}).
\end{aligned}
\end{equation}
Posterior distributions of atom features and coordinates are simple to compute as they are derived from normal distributions for continuous data (see Eq~\ref{eq:posterior.atom}). Here, we compute the posterior distribution for edge features,
\begin{equation}
\begin{aligned}
    p(\mathbf{{E}}_{t-1} | \mathbf{{E}}_{t}) & = \prod_{(i,j)\in \mathbf{E}} p(\mathbf{{e}}_{{t-1}_{ij}} | \mathbf{{e}}_{{t}_{ij}}) \\
    p(\mathbf{{e}}_{{t-1}_{ij}} | \mathbf{{e}}_{{t}_{ij}}) 
    & = \sum_{\mathbf{\hat{e}}_{ij} \in \mathbf{\hat{E}}} p(\mathbf{{e}}_{{t-1}_{ij}} | \mathbf{{e}}_{{t}_{ij}}, \mathbf{\hat{e}}_{ij}) p(\mathbf{\hat{e}}_{ij} | \mathbf{{e}}_{{t}_{ij}}) \\
    & = \sum_{\mathbf{\hat{e}}_{ij} \in \mathbf{\hat{E}}} p(\mathbf{{e}}_{{t-1}_{ij}} | \mathbf{{e}}_{{t}_{ij}}, \mathbf{\hat{e}}_{ij}) p(\mathbf{\hat{e}}_{ij}),
\end{aligned}
\end{equation}
where we choose
\[
    p(\mathbf{{e}}_{{t-1}_{ij}}| \mathbf{{e}}_{{t}_{ij}}, \mathbf{\hat{e}}_{ij}) = 
\begin{dcases}
     q(\mathbf{{e}}_{{t-1}_{ij}} | \mathbf{{e}}_{{t}_{ij}}, \mathbf{\hat{e}}_{ij}),  & \text{if } q(\mathbf{{e}}_{{t}_{ij}} | \mathbf{\hat{e}}_{ij}) > 0\\
    0,              & \text{otherwise.}
\end{dcases}
\]
The posterior distribution for discrete objects is given in Sec~\ref{sec:diffusion.processes}, as $q(\mathbf{{E}}_{t-1}| \mathbf{{E}}_{t}, \mathbf{{E}})$, but since the clean edge features $\mathbf{E}$ are unknown during sampling, we substitute it with the network approximation $\mathbf{\hat{E}}$, resulting in the posterior distribution $q(\mathbf{{E}}_{t-1}| \mathbf{{E}}_{t}, \mathbf{\hat{E}})$.

\section{Zeroth Likelihood Estimation}
\label{app:zeroth.likelihood}
For simplicity, we use $\mathbf{M}_t=(\mathbf{H_t,E_t,X_t})$ to denote the noisy molecule $\mathbf{\Tilde{M}}_t=(\mathbf{\Tilde{H}}_t,\mathbf{\Tilde{E}}_t,\mathbf{\Tilde{X}}_t)$ at time $t$.
The zeroth likelihood term for edge features is computed simply, which is just defined as the probabilities of the estimate of clean edge features computed from $\mathbf{E}_0$.

The zeroth likelihood term for atom positions can be computed similarly to the way it is defined in Eq~\ref{eq:posterior.atom},
\begin{equation}
\begin{aligned}
& p(\mathbf{X}|\mathbf{X}_0) = \mathcal{N}(\mathbf{X}|\frac{1}{\alpha_0}\mathbf{X}_0 -\frac{\sigma_0}{\alpha_0}\bm{\hat{\epsilon}}^0_{\mathbf{X}}, \frac{\sigma^2_0}{\alpha^2_0}\mathbf{I}) \\
& \mathbf{{X}}=\frac{1}{\alpha_{0}}\mathbf{{X}}_{0}-\frac{\sigma_{0}}{\alpha_{0}} \mathbf{\hat{{\bm{\epsilon}}}}^0_{\mathbf{X}} + \frac{\sigma_0}{\alpha_0}\bm{\epsilon}_{\mathbf{X}},
\end{aligned}
\end{equation}
and the reconstruction loglikelihood follows,
\begin{equation}
\begin{aligned}
    \log p(\mathbf{X}|\mathbf{X}_0) = \frac{1}{2} \mathbb{E}_{{\bm{\epsilon}}_{\mathbf{X}}^0 \sim N_{\mathbf{X}}(\mathbf{0, I})}\left[ \omega(0) \| {\bm{\epsilon}}_{\mathbf{X}}^0-\mathbf{\hat{{\bm{\epsilon}}}}^0_{\mathbf{X}} \|^2\right],
\end{aligned}
\end{equation}
where $\omega(0)= -1$.

The zeroth likelihood term for atom features is not easy to compute as atom features can be continuous and categorical. 
We follow \cite{hoogeboom2022equivariant} to compute the zeroth likelihood term for atom features. For integer molecular properties, the likelihood follows
\begin{equation}
\begin{aligned}
    p(\mathbf{H}|\mathbf{H}_0) & =\int_{\mathbf{H}-\frac{1}{2}}^{\mathbf{H}+\frac{1}{2}}\mathcal{N}(\mathbf{h}|\mathbf{H}_0, \sigma_0) \text{d}\mathbf{h} \\
    & = f_\text{CDF}(\frac{\mathbf{H}+\frac{1}{2}-\mathbf{H}_0}{\sigma_0})- f_\text{CDF}(\frac{\mathbf{H}-\frac{1}{2}-\mathbf{H}_0}{\sigma_0})
\end{aligned}
\end{equation}
where $f_{\text{CDF}}(\cdot)$ denotes the cumulative distribution function of a standard normal distribution. For categorical features like atom types, a one-hot encoding is applied and the likelihood is calculated as,
\begin{equation}
\begin{aligned}
    p(\mathbf{H}|\mathbf{H}_0) = \mathcal{C}(\mathbf{H}|\mathbf{p}), \quad \mathbf{p} \propto \int_{\mathbf{1}-\frac{1}{2}}^{\mathbf{1}+\frac{1}{2}} \mathcal{N}(\mathbf{h}|\mathbf{H}_0, \sigma_0) \text{d}\mathbf{h},
\end{aligned}
\end{equation}
where $\mathbf{p}$ is normalized to one and $\mathcal{C}$ is the categorical distribution, as suggested in \cite{hoogeboom2022equivariant}.

\section{Node2Graph \& Edge2Graph Functions}
\label{app:n2g.e2g}
Node2Graph $f_{\text{Node2Graph}}(\cdot)$ and Edge2Graph $f_{\text{Edge2Graph}}(\cdot)$ functions map node- and edge-level features to graph-level features, respectively.

The Node2Graph function transforms the node features $\mathbf{H} \in \mathbb{R}^{n\times F_{\text{in}}}$ by computing the mean, max, and min values for each node, then concatenating them and applying a linear transformation with weight matrix $W_{\text{Node2Graph}}$ and bias $b_{\text{Node2Graph}}$,
\begin{equation}
\begin{aligned}
    & \mathbf{H}_{\text{mean}} = \text{mean}(\mathbf{H}) \in \mathbb{R}^{1\times F_{\text{in}}} \\
    & \mathbf{H}_{\text{max}} = \text{max}(\mathbf{H})  \\
    & \mathbf{H}_{\text{min}} = \text{min}(\mathbf{H}) \\
    & \mathbf{H}_{\text{out}} = W_{\text{Node2Graph}}([\mathbf{H}_{\text{mean}}, \mathbf{H}_{\text{max}}, \mathbf{H}_{\text{min}}]) + b_{\text{Node2Graph}} \in \mathbb{R}^{1\times F_{\text{out}}}.
\end{aligned}
\end{equation}

The Edge2Graph function transforms the node features $\mathbf{E}\in \mathbb{R}^{n\times n \times F_{\text{in}}}$ by computing the mean, max, and min values for each node pair $(i, j)$, then concatenating them and applying a linear transformation with weight matrix $W_{\text{Edge2Graph}}$ and bias $b_{\text{Edge2Graph}}$,
\begin{equation}
\begin{aligned}
    & \mathbf{E}_{\text{mean}} = \text{mean}(\mathbf{E}) \in \mathbb{R}^{1\times 1 \times F_{\text{in}}} \\
    & \mathbf{E}_{\text{max}} = \text{max}(\mathbf{E}) \\
    & \mathbf{E}_{\text{min}} = \text{min}(\mathbf{E}) \\
    & \mathbf{E}_{\text{out}} = W_{\text{Edge2Graph}}([\mathbf{E}_{\text{mean}}, \mathbf{E}_{\text{max}}, \mathbf{E}_{\text{min}}]) + b_{\text{Edge2Graph}} \mathbb{R}^{1\times 1 \times F_{\text{out}}}.
\end{aligned}
\end{equation}

\section{Molecule Generation}
\label{app:molecule.generation}
\paragraph{Discussion on Metrics}
Following the methodology outlined in \cite{satorras2021n, hoogeboom2022equivariant}, we evaluate the atom and molecule stability, as well as the validity and uniqueness of the generated samples by building a molecule with RdKit and attempting to obtain a valid SMILES string from it. However, as seen in Table~\ref{tab:cond.generation.stable} and Table~\ref{tab:cond.generation.valid}, the statistics of the QM9 dataset are not perfect (not reaching 100\%). This is due to the limitations of the method used by RDKit to process molecules, as explained in \cite{hoogeboom2022equivariant}. RDKit first builds a molecule that contains only heavy atoms, then adds hydrogens to each heavy atom in a way that matches the valency of each atom to its atom type. As a result, invalid molecules mostly appear when an atom has a valency bigger than expected. Additionally, as stated in \cite{jo2022score}, this method is not perfect as the QM9 dataset contains charged molecules that would be considered invalid by this method.

Regarding novelty, \cite{vignac2021top} argue that the QM9 dataset is a comprehensive enumeration of small molecules that meet a set of specified constraints. Therefore, a novel molecule would not satisfy at least one of these constraints, indicating that the model does not accurately capture the correct data distribution. This makes evaluating and reporting novelty for molecules generated from the QM9 dataset difficult and potentially misleading.

\paragraph{Training Procedure} Our MUDiff model is trained using NVIDIA A100 GPUs. The model architecture consists of 6 layers, with 256-dimensional embeddings for atom and edge-level features, and 64-dimensional embeddings for graph-level features. Additionally, we use 8 attention heads, 300 feedforward dimensions for atom and edge-level features, 100 feedforward dimensions for graph-level features, and 0.3 dropout rate for all latent embeddings and attention values. The activation function used is SiLU, the learning rate is set to 1e-4, and weight decay is set to 5e-5. The optimizer used is Adam. These hyperparameter settings are chosen to balance the trade-off between computational cost and model performance. 

For the diffusion process, we use 1000 time steps and a cosine schedule for the diffusion coefficient over 10000 training epochs. This schedule is introduced in \cite{nichol2021improved}, where the coefficient is defined as $\alpha_t = \cos (0.5\pi(\frac{t}{T}+s)/(1+s))^2$, where $T$ is the total number of time steps and $s$ is a small value (e.g., $10^{-6}$). This schedule helps to ensure a smooth transition from an initial low diffusion coefficient at the start of the process to a high diffusion coefficient at the end. This schedule also helps avoid the problem of over-diffusion at the early stages of the process and under-diffusion at the later stages.

\section{Disscussion on Limitation \& Future Work}
\label{app:discussion}

\subsection{Scalability Issue}
\label{app:discussion.scale.compu}
\paragraph{Problem} Generating molecular structures using graph models presents a challenge in representing the structures in a way that can be processed by the model. One commonly used approach is to represent the structure as a dense adjacency tensor, where each element in the tensor corresponds to the presence or absence of a bond between two atoms. However, generating these dense tensors can be computationally expensive, particularly for larger or more complex molecular structures.

\paragraph{Approach \& Limitation} In our model, which includes a transformer and diffusion model, we generate 2D molecular structures and edge features by creating a dense adjacency tensor of size $n\times n\times b$, where n represents the number of atoms in a molecule and b represents the number of edge types. In the case of molecule scenarios, our model predicts a dense tensor of size $n\times n\times 4$, which includes four bond types (no-bond, single bond, double bond, and triple bond). One advantage of using dense tensors is that they can capture more detailed information about the molecular structure, including the precise location and type of each bond. This level of detail can be particularly crucial in cases where subtle differences in the structure can have a significant impact on the molecule's properties or behavior. However, a significant disadvantage of dense tensors is the computational cost required to generate and process them, which can limit the scalability and efficiency of the model.

\paragraph{Sparse Tensor Solution}
Sparse tensors can be a useful approach to reducing the computational cost of generating molecular structures. Sparse tensors are similar to dense tensors, but they only store the non-zero elements of the tensor, rather than the entire tensor.
To use sparse tensors for the problem of generating molecular structures, one approach is to represent the adjacency matrix as a sparse tensor. Rather than creating a dense tensor of size $n\times n\times b$, we can instead create a sparse tensor that only stores the non-zero elements of the adjacency matrix. 
To make predictions for sparse tensors, we can use specialized algorithms designed for sparse tensors. One common approach is to use sparse matrix multiplication algorithms, which can efficiently perform matrix operations on sparse tensors. These algorithms are designed to take advantage of the sparsity of the tensor to perform the required computations more efficiently.

\paragraph{Multi-resolution Representation Solution}
Multi-resolution representation of molecules is an approach to represent molecular structures at multiple levels of detail, allowing for more efficient processing while still capturing important structural information. Essentially, the idea is to represent the molecule in different ways, with each representation capturing different levels of detail.
One common approach is to use a hierarchical representation, where the molecular structure is represented as a series of nested substructures. For example, the molecule could be represented as a set of atoms, each associated with a set of neighboring atoms. This set of neighboring atoms could then be recursively expanded to include their own neighboring atoms, resulting in a hierarchical representation of the molecule that captures different levels of detail at different scales.
Another approach is to use a multi-scale representation, where the molecular structure is represented at different levels of detail using different feature maps or descriptors. For example, the molecule could be represented as a set of atoms, each associated with a descriptor that captures its physical properties.
By using multi-resolution representations, we can reduce the computational cost of generating molecular structures while still capturing important structural information.

\end{document}